\documentclass[letterpaper, 10 pt, conference]{ieeeconf}  %
\usepackage[load-configurations = abbreviations]{siunitx}
\usepackage{amsmath}
\usepackage{amsfonts}
\usepackage{graphicx}
\usepackage{subcaption}
\usepackage{booktabs}
\usepackage{cite}

\makeatletter
\let\NAT@parse\undefined
\makeatother

\usepackage{hyperref}

\graphicspath{ {./images/} }

\hypersetup{
    colorlinks=true,
    linkcolor=blue,
    urlcolor=blue,
    citecolor=black,
    }

\IEEEoverridecommandlockouts                              %

\title{\LARGE \bf
Learning Coordinated Terrain-Adaptive Locomotion by Imitating a Centroidal Dynamics Planner
}
\author{Phil\'emon Brakel, Steven Bohez, Leonard Hasenclever, Nicolas Heess, Konstantinos Bousmalis %
\thanks{Deepmind London,}
\thanks{\texttt{\{philemon,sbohez,leonardh,heess,konstantinos\}}}
\thanks{\texttt{@deepmind.com}}
\thanks{Video: \url{https://youtu.be/dvFwkvef2Ww}}}

\begin{document}

\maketitle
\thispagestyle{empty}
\pagestyle{empty}

\begin{abstract}
Dynamic quadruped locomotion over challenging terrains with precise foot placements is a hard problem for both optimal control methods and Reinforcement Learning (RL).
Non-linear solvers can produce coordinated constraint satisfying motions, but often take too long to converge for online application. RL methods can learn dynamic reactive controllers but require carefully tuned shaping rewards to produce good gaits and can have trouble discovering precise coordinated movements. Imitation learning circumvents this problem and has been used with motion capture data to extract quadruped gaits for flat terrains. However, it would be costly to acquire motion capture data for a very large variety of terrains with height differences. In this work, we combine the advantages of trajectory optimization and learning methods and show that terrain adaptive controllers can be obtained by training policies to imitate trajectories that have been planned over procedural terrains by a non-linear solver.
We show that the learned policies transfer to unseen terrains and can be fine-tuned to dynamically traverse challenging terrains that require precise foot placements and are very hard to solve with standard RL.

\end{abstract}

\section{Introduction}

Robots with legs can access locations that are out of reach for platforms with wheels and can carry loads that would be infeasible for airborne vehicles. This makes legged locomotion an application area with practical impact. On top of that, it is also a challenging research domain with complex control and perception problems for which we know that solutions exist in nature.

While recent hardware and algorithmic developments have led to impressive legged locomotion demonstrations of both learned controllers \cite{Lee2020,Siekmann2021blind}, model-based planning approaches \cite{kalakrishnan2010fast,bellicoso2018dynamic} and combinations of the two \cite{Tsounis2020deepgait,Carius2020mpc}, there appears to be a trade-off between the overall agility of the systems and the ability to traverse complex terrains that require precise foot placements. Broadly speaking, methods that involve planning tend to struggle more with dynamic movements while learned controllers struggle more with producing precise coordinated foot placements and desirable gaits.

For many planning-based locomotion controllers, the lack of dynamic movements is attributable to the necessity for the planner to provide trajectory plans fast enough to run online. This requirement often rules out general non-linear solvers and forces the search space to be constrained to more conservative motions or the foothold planning and the dynamics planning to happen separately. When footholds are planned independently of the robot dynamics, they also have to be more conservative to guarantee feasibility, especially on rough terrains.

Reinforcement learning (RL) methods generate closed-loop control policies which can produce very dynamic and robust gaits that have been successfully transferred to hardware \cite{Siekmann2021blind,Hwangbo2019learning}. However, it often requires significant tuning and the use of many shaping rewards to produce desirable gaits.
For trajectory optimization this is less of an issue as constraints can easily be added and tuned relatively quickly.
Furthermore, it can be hard for learning algorithms to discover very precise movements over long time horizons via random exploration. For these reasons, many successful learning methods employ planning methods to provide more effective exploration \cite{lowrey2018plan,levine2013,Carius2020mpc}. A downside of such hybrid approaches is that even if the planner doesn't need to run online during hardware deployment, the rate at which it can generate trajectories can quickly become a bottleneck in the learning iteration. This still limits most use cases of planners in learning systems to conservative trajectory spaces or relatively short time horizons compared to state-of-the-art trajectory optimization.

One way to circumvent exploration problems and also prevent undesirable gaits, is through imitation learning. Motion capture data from a dog has been shown to be a useful source of behaviors from which natural looking gaits were successfully transferred to a quadruped robot \cite{peng2020animals}. However, obtaining motion capture data for a wide variety of terrains would be very costly. It is also not clear how much data would be needed and whether imitation learning provides the required generalization to traverse unseen terrains.

In this paper, we demonstrate that imitation learning from planner data can be used to efficiently create general terrain-adaptive controllers with precise foot placements.
We first use a non-linear centroidal dynamics planning formulation in an offline fashion to generate a dataset of trajectories that traverse a diverse set of procedurally generated terrains. Subsequently, we use reinforcement learning to obtain policies that imitate these trajectories. This setup has two advantages over conventional optimal control and RL methods: first, the trajectory optimizer can take as much time as is needed to find long time horizon trajectories over complex terrains because it doesn't need to operate in real time; second, the reinforcement learning agent does not need to discover the coordinated behaviors and constrained gaits through exploration with shaping rewards.
Concretely, we make the following contributions:
\begin{enumerate}
    \item We provide a recipe for generating varied terrain traversal datasets using a centroidal dynamics planner.
    \item We demonstrate how to use RL to effectively imitate the generated trajectories.
    \item We show empirically that the learned policies display reasonable zero-shot performance on unseen terrains and that they can be fine-tuned to traverse complex stepping stones and mixed procedural terrains that are hard to learn from scratch.
\end{enumerate}

\section{Related Work}

\subsection{RL for Terrain-Aware Locomotion}
RL has been used to learn terrain-adaptive policies for various simulated characters/robots \cite{Song2019v,heess2017emergence}. In recent work \cite{escontrela2020zero}, a multi-task setup was used to learn a quadruped controller that was able to traverse stairs and various other rough terrains in simulation. This controller used a pattern generator modulating action space and it is not clear if the system could learn the precise movements needed for the environments in our work.
Similarly, in work concurrent to ours, evolved pattern generators were combined with RL to traverse challenging terrains \cite{shi2021reinforcement}.
In the ALLSTEPS approach \cite{xie2020allsteps}, RL was used to train policies that were able to walk over stepping stones requiring precise foot placements. The behaviors were learned using a curriculum, highly task specific rewards and feature observations based on step-by-step target foothold locations.
Other types of curricula have been used together with shaping rewards to learn blind \cite{Lee2020} and very recently also depth perceptive \cite{rudin2021learning} terrain adaptive controllers via RL. It is not clear whether such curricula can easily be designed for any type of terrain.

\subsection{Combining RL and Planning for Locomotion}
Various systems combine ideas from RL and optimal control for locomotion.
DeepGait \cite{Tsounis2020deepgait}, for example, is a hierarchical system in which a higher level policy produces goal poses that are judged for feasibility by a centroidal dynamics planner. This system can traverse challenging terrains but its motions tend to be conservative and fairly static. This is probably due to the simplified (convex) formulation that was used to make the planner fast enough.
In MPC-Net \cite{Carius2020mpc}, trajectories are generated using an MPC controller and optimized with a loss that minimizes the original control Hamiltonian. One advantage of this method is that it takes the original optimal control constraints into account. However, the learned policy is limited by what the teacher MPC controller can do and the teacher still needs to employ an optimal control solver in its inner-loop.
RLOC \cite{Gangapurwala2020rloc} combines RL and optimal control by learning a terrain-aware foothold planner that interfaces with a whole-body controller (WBC). This systems was shown to traverse a wide variety of terrains. Unlike our approach, RLOC still uses a WBC which may limit the types of movements that can be selected by the perceptive foothold setting policy. 
GLiDE \cite{xie2021glide} learns an RL policy for the centroidal dynamics directly. This system was shown to traverse narrow beams and small stepping stones. However, the foothold planning was done with terrain specific hand-designed heuristics.
Pre-planned trajectories have been used to guide learning of policies \cite{gangapurwala2020guided} to aid exploration and constraint satisfaction but so far this method has only been applied to flat terrains.
Finally, learning methods have also been combined with motion planning to speed up the planning process by training neural networks to provide good initializations for the solver \cite{Melon2020reliable,kwon2020fast}. The obtained plans still need to be executed by a WBC but such methods are complementary to our approach and could be used to speed up the dataset construction.

\subsection{Planner Imitation for Legged Robots}
In very recent work \cite{bogdanovic2021model} that is perhaps most similar to ours, a single trajectory was generated with a motion planner and RL was used to imitate it in a timing independent way and learn a bounding or hopping behavior which was fine-tuned and evaluated on a robot. Our work is different in that we generate an entire dataset of trajectories and use it for learning terrain adaptive policies that can generalize and be fine-tuned for a variety of challenging terrains.

\subsection{Character Animation and Mocap Tracking}
Our work is inspired by recent methods for animated character control via RL-based imitation of motion capture data \cite{Peng2018deepmimic,Merel2018neural,Hasenclever2020comic}. %
Motion capture data has also been used to transfer dog behaviors to a real quadruped platform \cite{peng2020animals} or to provide adversarially learned motion priors \cite{peng2021amp}.
Holden et al.\ \cite{Holden2017phase} use terrains that were fitted to existing motion capture trajectories for kinematic replay. This is similar in spirit to our method for distorting terrains but used in a different context.

\section{Method}

\begin{figure*}
    \centering
    \includegraphics{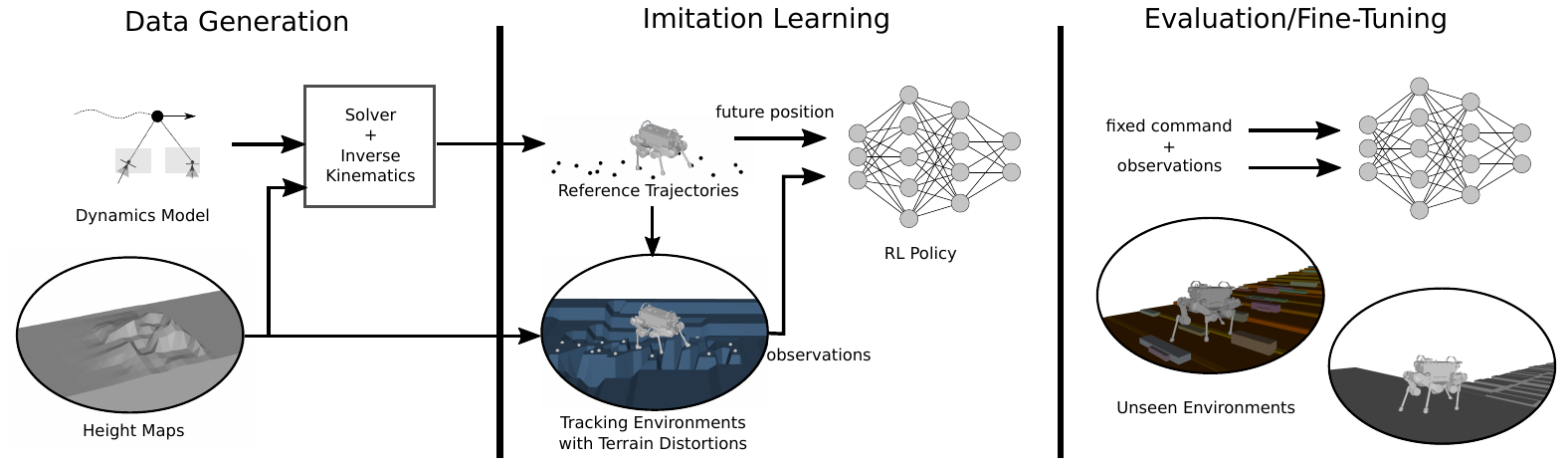}
    \caption{A schematic overview of our method that can be separated in a \emph{data generation} step in which the terrains and planned trajectories are generated, an \emph{imitation learning} step in which the terrains are distorted and policies are trained to imitate the trajectories and an \emph{evaluation/fine-tuning} step where the input commands are changed to produce straight walking behavior and there is potential further training for new types of terrains.}
    \label{fig:overview}
\end{figure*}

As Fig. \ref{fig:overview} shows, our method can be divided into three phases: data generation, imitation learning and evaluation/fine-tuning. To generate the data, we first generated procedural terrains as detailed in Sec.\ \ref{sec:terrains}, and defined a feasible planner formulation which is described in Sec.\ \ref{sec:formulation}. In Sec.\ \ref{sec:generation}, we detail how this formulation was used to produce diverse trajectories that traverse the terrains. To explain how imitation learning was used to reproduce the trajectories with a feedback policy, we describe the RL tracking task in Sec.\ \ref{sec:tracking}, the learning algorithm in Sec.\ \ref{sec:learning}, the observations and network architecture in Sec.\ \ref{sec:observations} and finally the terrain distortions in Sec.\ \ref{sec:distortions}.
In Sec.\ \ref{sec:tuning} we describe how evaluation and fine-tuning were done for new terrains.

\subsection{Procedural Terrains}
\label{sec:terrains}
\begin{figure}
    \centering
    \includegraphics[scale=0.12]{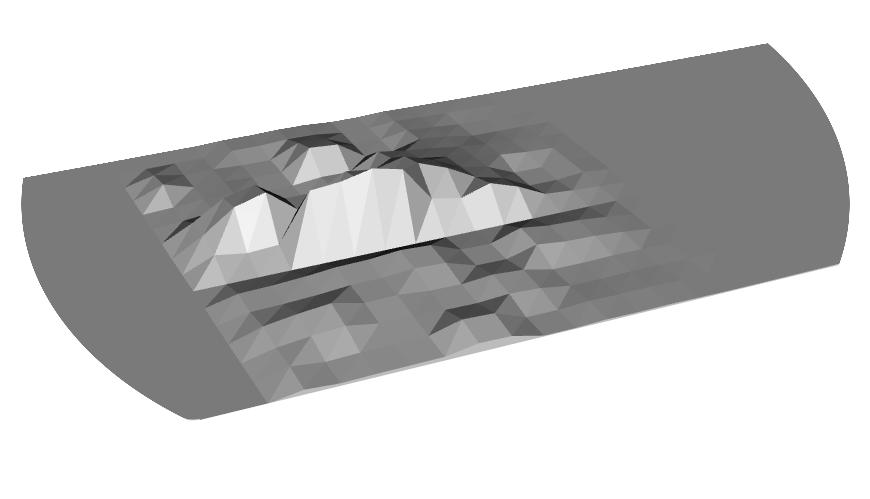}
    \includegraphics[scale=0.12]{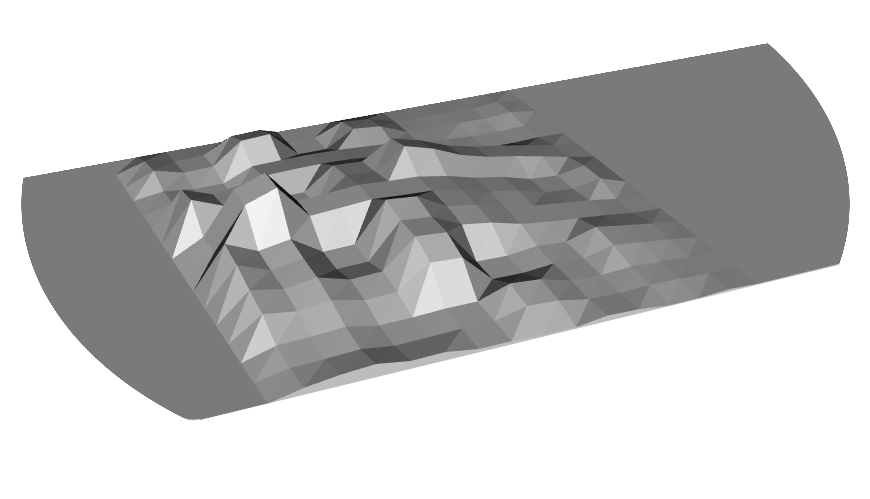} 
    \includegraphics[scale=0.12]{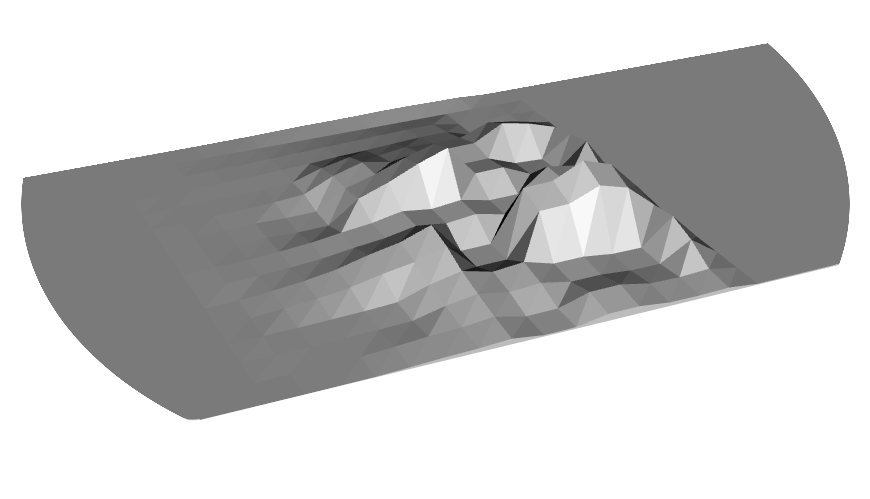}
    \caption{Examples of procedural terrains provided to the planner and used in some of the evaluation environments.}
    \label{fig:heightfields}
\end{figure}

To obtain a diverse set of trajectories with variations in contact positions and orientations, we implemented a system for procedural terrain generation.
We represented elevation data as a triangulated grid. %
We aimed to have terrains with both flat and tilted surfaces. To do this, we created random grids of tiles with differing heights and subsequently scaled these terrains by multiplying with Gaussian functions representing round bumps.
The height fields were scaled down to always have a maximum height of $\SI{30}{cm}$.
See Fig.\ \ref{fig:heightfields} for examples of height fields generated with this procedure and Appendix \ref{app:terrain} for a more detailed description.

\subsection{Planner Formulation}
\label{sec:formulation}
Our motion planner formulation extends the centroidal dynamics planner proposed by Winkler et al.\ \cite{Winkler2018-zl} and is implemented on top of their open-source library (TOWR) which also provides an interface for the IPOpt solver \cite{Wachter2006} that we used for optimization. The default formulation in TOWR is a nonlinear program (NLP) of the \emph{simultaneous} type where both the states and controls are optimization variables and the dynamics are represented as constraints. The dynamics constraints describe the motion of a single rigid body (the center of mass of the robot) and are a function of the contact forces and paths of the end-effectors (feet). The motions and paths are represented as splines of cubic Hermite polynomials. Optimization is done over the parameters of these polynomials, which can be interpreted as the positions and velocities of the spline nodes and the duration of each spline segment.

More precisely, the formulation defines a search over the CoM positions $\mathbf{r}(t)\in \mathbb{R}^3$, the base orientations $\mathbf{\theta}(t)\in \mathbb{R}^3$, the end-effector positions $\mathbf{p}_i\in \mathbb{R}^3$, force profiles $\mathbf{f}_i$,  and the phase durations for each of the end-effectors $\Delta T_{i,j}$, where $i$ indexes the end-effector and $j$ the phase.
The forces are only non-zero during the contact phases.
The optimization has a trivial cost but the solution must satisfy a set of constraints that can be grouped in the following way:
\begin{itemize}
    \item Dynamics of the single body system enforced at fixed time intervals.
    \item Contact force constraints to ensure positive and bounded values in the terrain surface normal direction and adherence to the friction pyramids.
    \item Swing height constraints that ensure that the position spline nodes are above the terrain.
    \item Kinematic limit constraints that restrict the end-effector positions to bounding boxes at fixed offsets of the center of mass.
\end{itemize}
We refer to the work by Winkler et al.\ \cite{Winkler2018-zl} for the details of the formulation and will only elaborate on our changes.

Similar to previous work \cite{Melon2020reliable}, we modified the formulation to improve the physical feasibility of the optimized trajectories.
Since the force constraints are by default only enforced at the nodes of the end-effector force splines, it is possible for the end-effectors in contact phase to generate forces that are large or even negative in the normal direction in-between two node values.
Similarly, the swing height constraints on the end-effector positions are only enforced at the node positions and we observed end-effectors passing through the terrains.
To address this, we also enforced the force and swing height constraints at fixed time intervals ($\delta t=0.08s$ and $\delta t=0.04s$ respectively). These additional constraints made the optimization problem harder and we increased the number of nodes per spline segment for the end-effector position trajectories from 2 to 3 to compensate and provide additional flexibility.
This is different from the approach by Melon et al.\ \cite{Melon2020reliable}, where different constraints and also \emph{cost} terms where added to minimize ground-reaction forces and to encourage higher swing-phase foot positions. 
Since our goal was to generate a \emph{diverse} dataset and not to directly execute the trajectories with a WBC, our priorities were different from Melon et al.\ and we opted for a more permissive formulation without any cost terms.
We think that cost terms are more likely to reduce the diversity of the optimization results compared to a formulation where the optimization landscape is flat within the region where the constraints are satisfied.

\subsection{Data Generation}
\label{sec:generation}
To create a dataset of trajectories and terrains, the terrain generation routine from Section \ref{sec:terrains} and the trajectory solver are called repeatedly.
While the randomness of the terrains should lead to a diverse set of trajectories, we also added some randomization to the initialization of the solver. The end-effector and CoM spline node positions were initialized using the procedure from TOWR using one of the predefined gait patterns (in TOWR denoted as `C1', which defines a fly trot pattern\footnote{A fly trot is probably not the best gait for scaling complex terrains but it appeared to be a good initial setting for the solver.}), to which we added normally distributed noise with a standard deviation of $\sigma=\SI{10}{cm}$. Similarly, we added noise with a standard deviation of $\sigma=\SI{5}{N}$ to the initial values of the spline nodes of the force variables.
We set the goal location to be $\SI{2.3}{m}$ in front of the robot (based on the terrain size and optimization time) and set the amount of time allowed to reach the goal to $\SI{4.6}{s}$ to promote an average forward velocity of $\SI{0.5}{m/s}$. To give the solver enough time for difficult trajectories but to also prevent problematic optimization regions from halting progress, we set the maximum CPU time of the solver to 30 minutes. 
The result of the optimization are the splines describing the CoM and end-effector motions and the durations of the swing and stance phases. These were sampled into discrete values at a rate of $\SI{100}{Hz}$.

To compute joint states (which were used for initialization) from the end-effector and CoM trajectories, we used a geometric approach based on the method implemented in the ANYbotics codebase. To obtain a consistent mapping, we constrained the knees to always bend inwards. For the ANYmal B robot model that we used, this corresponds to the so called x-configuration of the legs.
See \cite{sen2017inverse} for an example of how to compute these quantities for a similar quadruped.

\begin{figure}
    \begin{subfigure}{0.20\textwidth}
        \includegraphics[width=\textwidth]{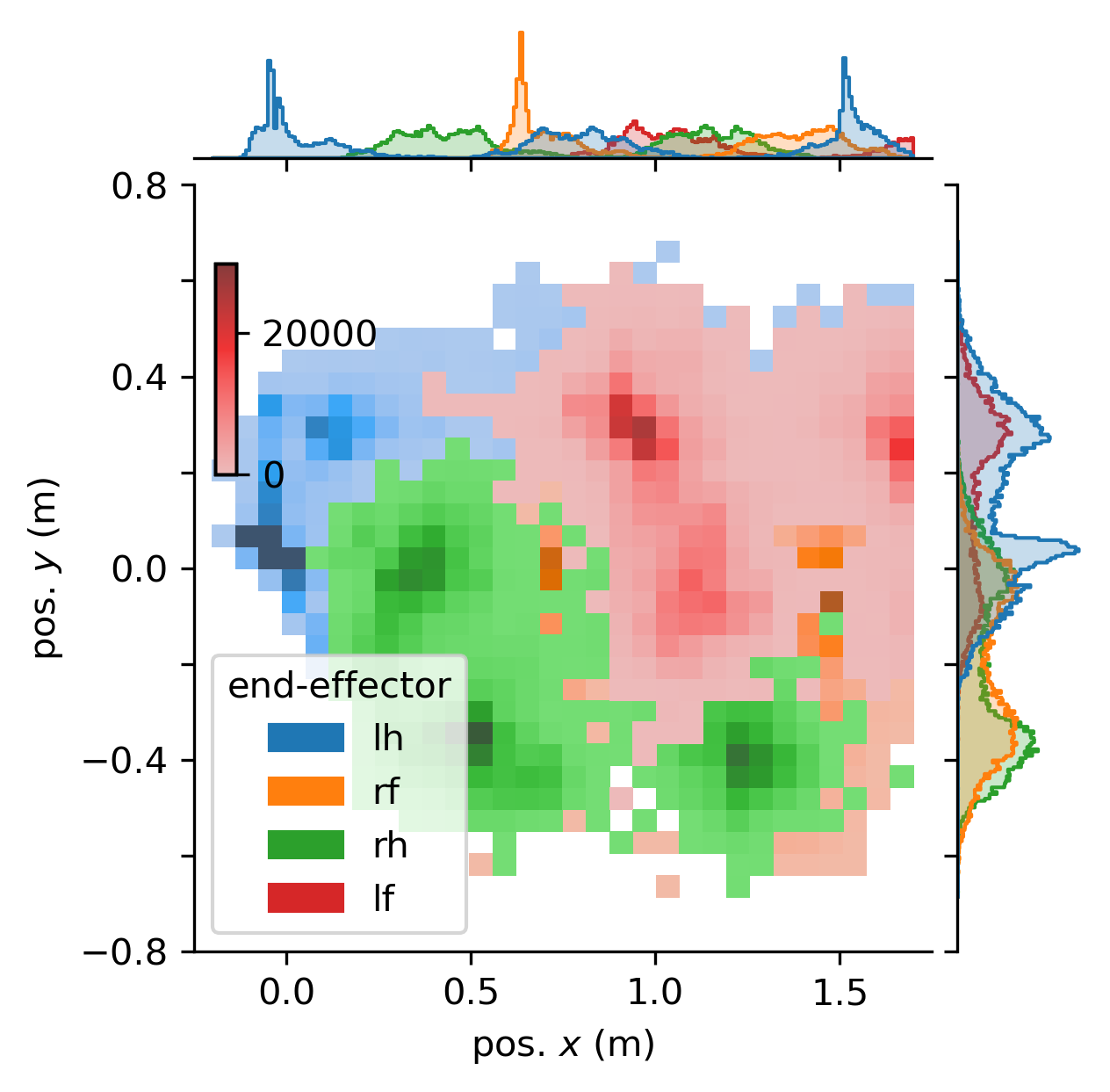}
        \caption{}
        \label{fig:eeshort}
    \end{subfigure}
    \begin{subfigure}{0.25\textwidth}
        \includegraphics[width=\textwidth]{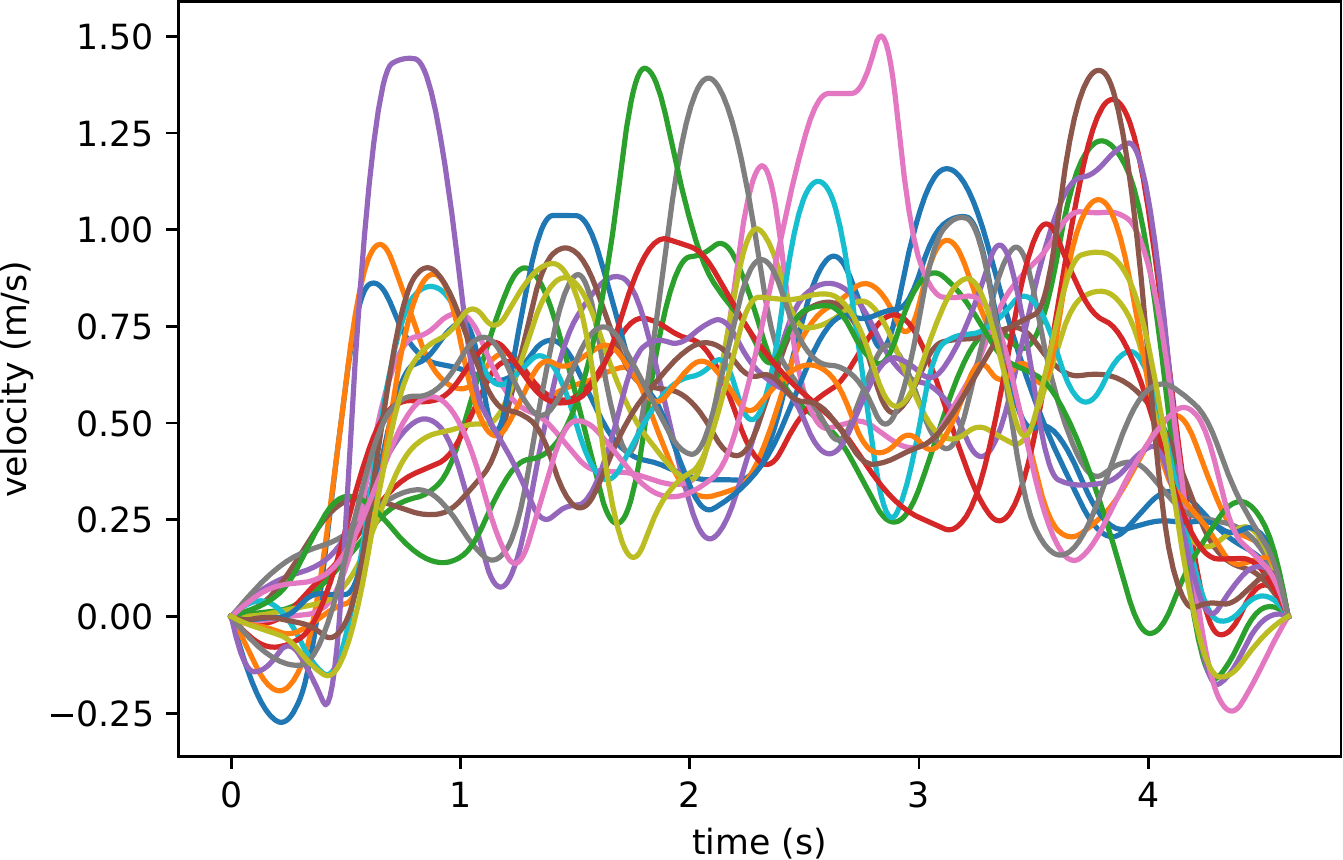}
        \caption{}
        \label{fig:velocityx}
    \end{subfigure}
    \caption{
    Distribution plots of the dataset. Fig.\ \ref{fig:eeshort} shows the planar end-effector contact distribution data in the world frame, where $x$ is aligned with the robot facing forward, up to $1.75$ meters (the end-position was always the same so we left it out of the distribution). Fig.\ \ref{fig:velocityx} shows the velocity of the center of mass projected on the $x$-axis for 20 randomly chosen trajectories as a function of time.}
    \label{fig:eedist}
\end{figure}

The final dataset has $8926$ clips of $\SI{4.6}{s}$ each, corresponding to about $11.4$ hours in total.
Each clip $C_n$ was represented by a tuple $(\mathcal{M}^n, \{\mathbf{p}^n_{\text{com}}(t), \mathbf{v}^n(t), \mathbf{\omega}^n(t),\mathbf{p}_i(t),g^n_i(t), \mathbf{q}(t),\mathbf{\dot{q}}(t)\colon 0\le t\le T\})$ combining a terrain image and discretized trajectory information, where $\mathcal{M}$ is the $16 \times 16$ terrain image, $\mathbf{p}^n_{\text{com}}(t)$ the center-of-mass position, $\mathbf{v}^n(t)$ the linear CoM velocity, $\mathbf{\omega}^n(t)$ the angular CoM velocity, $\mathbf{p}_i(t)$ the end-effector positions, $g^n_i(t)$ is a binary indicator of whether each end-effector $i$ was in contact with the ground or not and $\mathbf{q}(t)$ and $\mathbf{\dot{q}}(t)$ are the joint positions and velocities obtained by inverse kinematics and finite differences.
Fig.\ \ref{fig:eeshort} shows the distribution of the end-effector contacts in the world frame $xy$-plane. The positions are clearly spread out and indicative of a diverse set of trajectories. The forward velocity of the trajectories varied significantly over time and among different trajectories as can be seen in Fig.\ \ref{fig:velocityx}.

\subsection{Tracking Task}
\label{sec:tracking}
To train a policy to track the planned trajectories, we used setup similar to the multi-clip motion capture tracking task from Hasenclever et al.\ \cite{Hasenclever2020comic}.
Each episode, a clip was chosen from the dataset and a random starting frame was selected to initialize the simulated robot state.
The episode ran until the end of the clip was reached or a termination criterion had been triggered. The agent was rewarded based on the similarity of the simulated trajectory to the reference clip.

We used the same truncation criterion as Hasenclever et al.\ \cite{Hasenclever2020comic} to terminate episodes when the robot would stray too far from the reference trajectory. This criterion is defined as
\begin{align}
\epsilon &= \frac{1}{3|\mathcal{B}|}\sum_{i\in\mathcal{B}}\|\mathbf{p}_{i} - \mathbf{p}^{\text{ref}}_{i}\|_{1} + \frac{1}{|\mathcal{J}|}\sum_{j\in\mathcal{J}}\|q_{j} - q^{\text{ref}}_{j}\|_{1}, \\
r_{\text{trunc}} &= 1 - \frac{\epsilon}{\tau},
\end{align}
where the vectors $\mathbf{p}_i$ represent the positions of all bodies and the values $q_j$ are the joint positions with indices $\mathcal{J}$. An episode was terminated when $r_{\text{trunc}}>\tau$, with $\tau=0.5$ for all our experiments.
Episodes were also terminated on contacts that were not between the end-effectors and the environment.

The tracking reward consists of multiple objectives.
The total reward used to monitor tracking performance consists of five terms defined as
\begin{align}
r_{\text{com}} &= 0.2\cdot\exp(-80\|\mathbf{p}_{\text{com}} - \mathbf{p}^{\text{ref}}_{\text{com}}\|^2), \label{eq:r_first} \\
r_{\text{ee}} &= 0.2\cdot\exp\left(-80\sum_i\|\mathbf{p}_{\text{ee,i}} - \mathbf{p}^{\text{ref}}_{\text{ee,i}}\|^2\right), \\
r_{\text{linvel}} &= 0.2\cdot\exp(-10\|\mathbf{\dot{p}}_{\text{com}} - \mathbf{\dot{p}}^{\text{ref}}_{\text{com}}\|^2), \\
r_{\text{angvel}} &= 0.2\cdot\exp(-10\|\mathbf{\omega}_{\text{com}} - \mathbf{\omega}^{\text{ref}}_{\text{com}}\|^2), \\
r_{\text{quat}} &= 0.2\cdot\exp(-2\|\mathbf{q}_{\text{com}} \ominus \mathbf{q}^{\text{ref}}_{\text{com}}\|^2)\label{eq:r_last},
\end{align}
where $\mathbf{p}_{\text{com}}$ and $\mathbf{p}_{\text{ee,i}}$ are the positions of the center of mass and the end-effectors, $\mathbf{\dot{p}}_{\text{com}}$ and $\mathbf{\omega}_{\text{com}}$ are the linear and angular velocities of the center of mass and $\mathbf{q}_{\text{com}}$ is the quaternion representing the base orientation at the center of mass.

\subsection{Learning Algorithm}
\label{sec:learning}
Our learning problem can be described as a multi-objective POMPD where there is more than one reward function and we have only access to observations rather than complete states.
To optimize the set of objectives, we used the Multi-Objective variant of the VMPO algorithm (MO-VMPO) \cite{Abdolmaleki2020distributional,Song2019v}, which has been shown to work well for motion capture tracking. MO-VMPO is an on-policy actor-critic policy improvement method, suitable for learning with neural networks. Like other actor-critic methods, there is a policy (actor) network $\pi(a_t | o_t)$ that predicts actions $a_t$ given observations $o_t$ and a value function network (critic) $V_l(o_t)$ which maps observations to estimates of the expected discounted future returns for each objective indexed by $l$. 
The MO-VMPO algorithm was configured to optimize the reward terms in Equations \ref{eq:r_first} to \ref{eq:r_last} separately using individual critic outputs for value estimation.
The policies were trained at a control rate of $\SI{100}{Hz}$, with a discount factor of $0.99$, using a batch size of $1024$, a trajectory unroll length of 20, MO-VMPO $\epsilon_k=0.002$ for the $k$ objectives and a learning rate of $10^{-4}$. Optimization of the networks was done with Adam \cite{kingma2014adam}.
See Appendix \ref{app:hypers} for more hyperparameter settings.

\subsection{Observations and Architecture}
\label{sec:observations}
\label{sec:obs}
At each time step, the policy observed a tuple of features
\begin{align}
\phi(t)=[&\mathbf{M}(t), \mathbf{q}(t),\mathbf{\dot{q}}(t),\mathbf{v}(t),\mathbf{\omega}(t), \mathbf{p}_i(t),h(t), \nonumber \\
&\mathbf{c}(t),\mathbf{X}(t),\mathbf{\hat{X}}(t), a_{t-1}, p^{y}_{\text{com}},\mathbf{z}(t)],
\end{align}
where $\mathbf{M}$ is an image with local height information ($32\times32$ pixels representing the $\SI{1.7}{m}\times \SI{1.7}{m}$ square centered $\SI{40}{cm}$ in front of the CoM, updated at $\SI{10}{Hz}$), $\mathbf{q}$ and $\mathbf{\dot{q}}$ are the positions and velocities of the joints, $\mathbf{v}$ and $\mathbf{\omega}$ the linear and angular velocities of the base, $\mathbf{p}$ the end-effector positions, $h$ the height of the base and $\mathbf{X}$ the orientation of the base represented as a rotation matrix. The features $\mathbf{c}$ and $\mathbf{\hat{X}}$ contain the relative position and orientation of the base with respect to the pose of the robot at the last time step that the local height image was updated. The previously taken action was observed as well. The absolute position of the center of mass on the $y$-axis $p^y_{\text{com}}$ was added to allow policies to track a straight path during evaluation. Finally, a position `command' $\mathbf{z}(t)$ was provided which consists of the pair $(p^{x,\text{ref}}_{\text{com}}(t+10)-p^x_{\text{com}}(t), p^{y,\text{ref}}_{\text{com}}(t+10))$, i.e., the relative $x$-coordinate of the reference center of mass and its absolute $y$-coordinate $10$ timesteps in the future. These commands were added to the observation to help the controller disambiguate between different trajectories to track. The choice to use absolute $y$-coordinates was again to easily switch to straight line following in downstream tasks by setting this $y$ command to zero.

\begin{figure}
    \centering
    \includegraphics[scale=0.32]{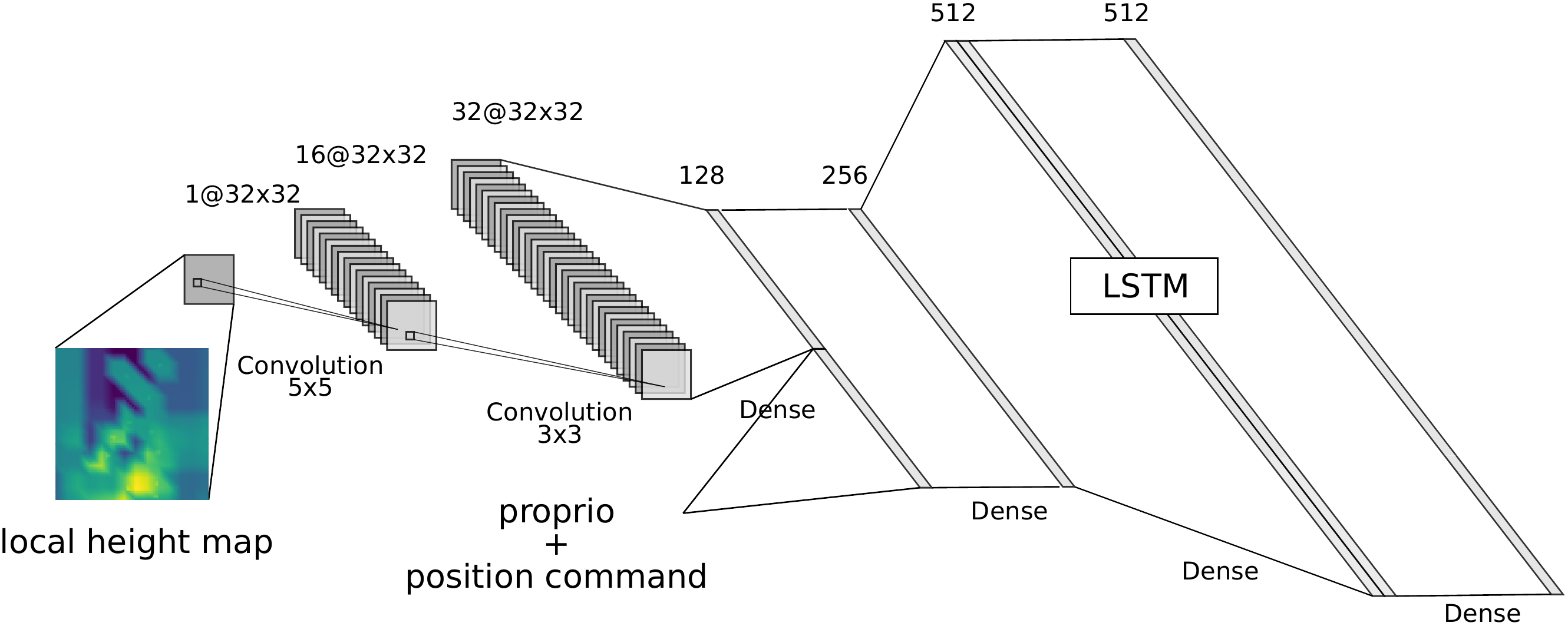}
    \caption{Network architecture of the actor network up to the penultimate layer. The final layer maps densely to the parameters of the Gaussian action distribution.}
    \label{fig:architecture}
\end{figure}

The actor is a conditional normal distribution parameterized by a deep neural network.
Since the observations don't provide full state information, we added memory to the network by incorporating an LSTM \cite{hochreiter1997long}.
Convolutional layers were used to process the local terrain information.
See Fig.\ \ref{fig:architecture} for a schematic representation of the actor network architecture.
The critic network received the same observations as the actor but also an embedding vector representing the clip number and a set of future reference body positions and quaternions.
See Appendix \ref{app:architecture} for more details about the networks architectures used.
Finally, the critic parameters were kept during fine-tuning but it's additional observations (see Sec. \ref{sec:obs}) were replaced with zeros. We didn't find that resetting the critic parameters lead to very different results (see Appendix \ref{app:reloading}).

\subsection{Terrain Distortions}
\label{sec:distortions}
\begin{figure}
    \centering
    \includegraphics[scale=0.09]{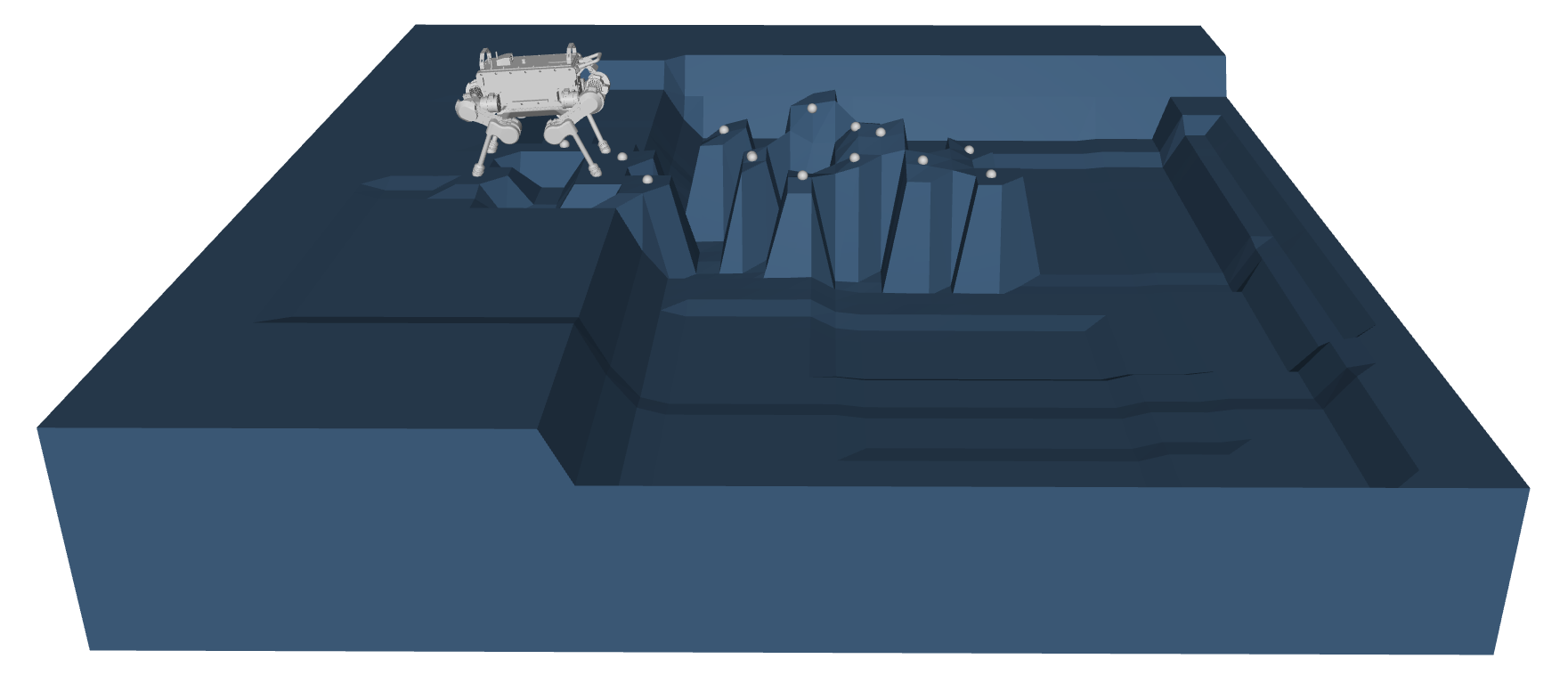}
    \includegraphics[scale=0.09]{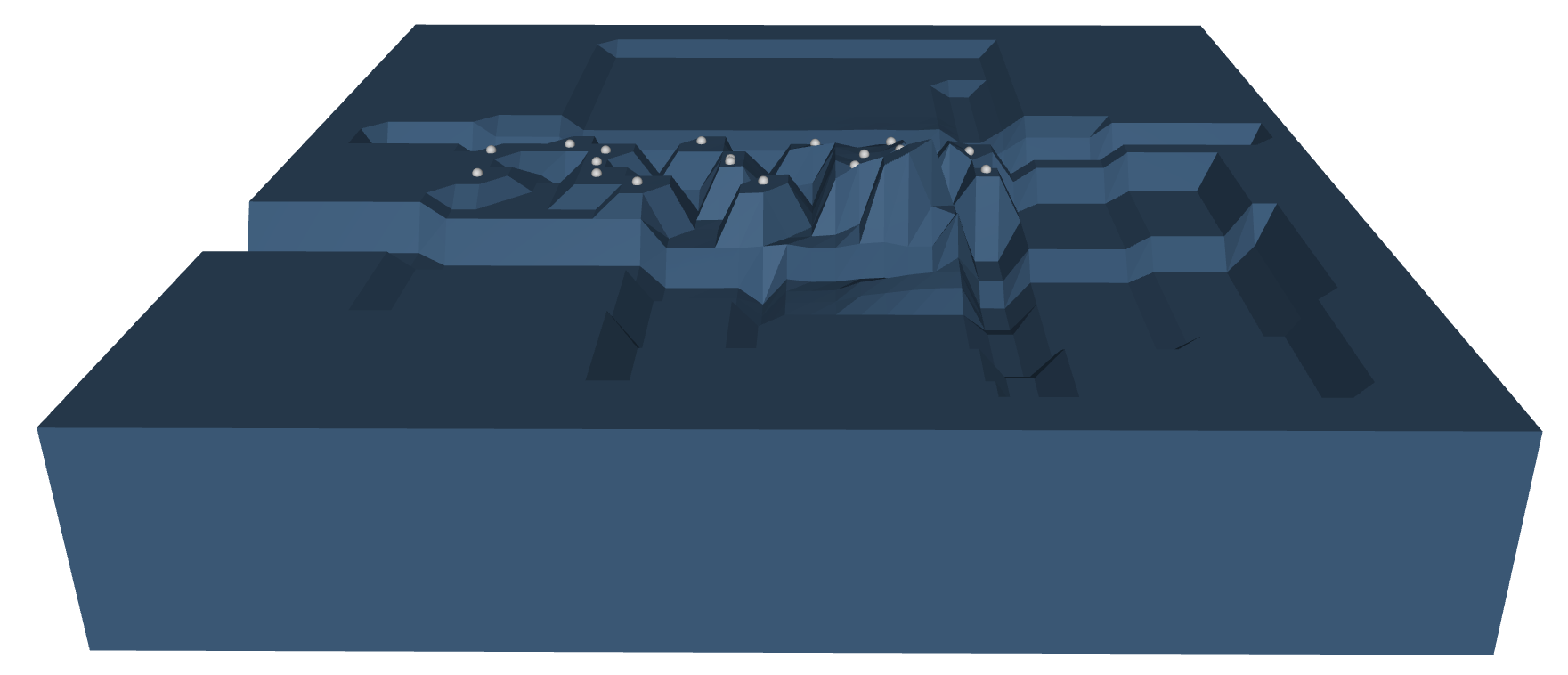}
    \includegraphics[scale=0.09]{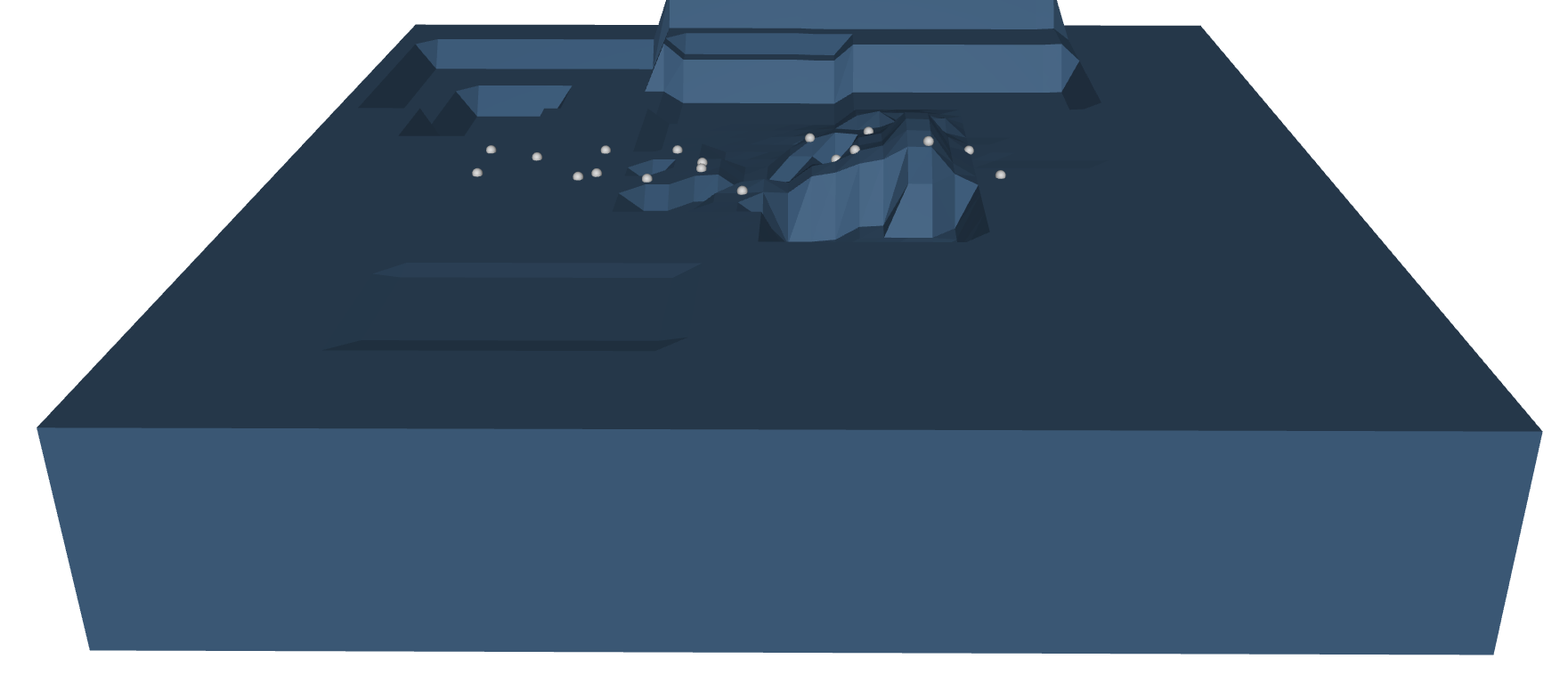}
    \includegraphics[scale=0.09]{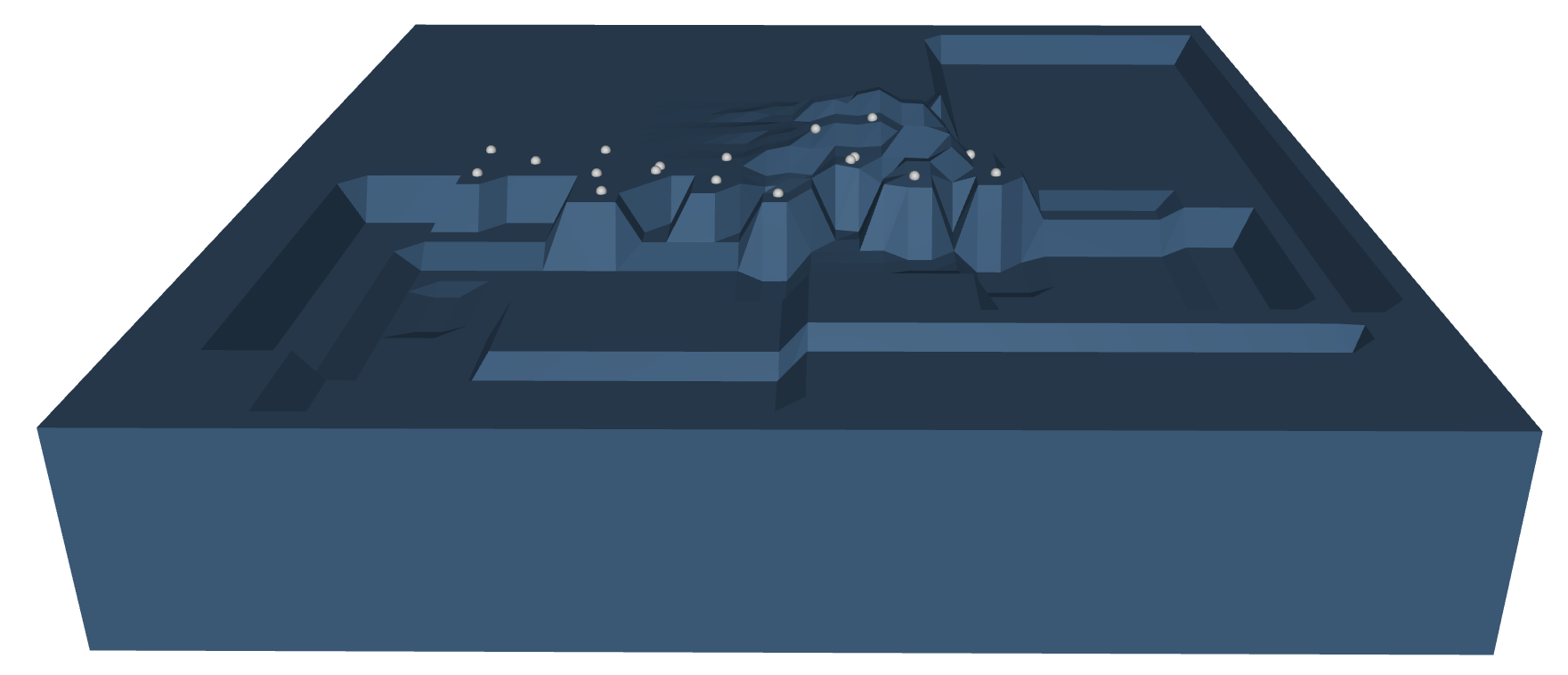}
    \caption{Examples of distorted terrains after embedding the height maps in a larger area and rescaling the height values at random rectangular regions.}
    \label{fig:distortions}
\end{figure}

For half the seeds of the tracking experiments, the terrains were \emph{distorted} to provide more diverse height map observations while respecting the locations and orientations of the planned end-effector contacts. The $16\times 16$ height map of a clip was first embedded in the center of a $46\times 46$ matrix. Subsequently 8 sets of 4 integers representing the corner indices of rectangles where sampled uniformly. The heights within sampled rectangles from the middle region with the original height map or the flat region in front of it were scaled with a factor sampled uniformly from $(0.7, 1.0)$. If the rectangle was in one of the other regions, the scaling was sampled from the interval $(0.8, 1.5)$. Finally, all terrain surface triangles overlapping with $\SI{10}{cm}$ squares centered on the end-effector contact locations were reset to their original heights.
See Fig.\ \ref{fig:distortions} for examples of distorted terrains.

\subsection{Evaluation and Fine-Tuning}
\label{sec:tuning}
During zero-shot evaluation and fine-tuning experiments, the position commands $\mathbf{z}(t)$ were fixed to $(0.05, 0.0)$.
During fine-tuning, the reward function was also adapted. The squared difference in the reward term $r_{\text{com}}$ was replaced with $(v_x - 0.5)^2 + (p^{y,\text{ref}}_{\text{com}})^2$ to promote straight walking at a fixed velocity and the other rewards terms were discarded.
This also means that no updates were done for the value heads of the critic representing these rewards and that these value heads were ignored in the MO-VMPO policy improvement step.
Note that we replaced a position target here with a velocity target but in a way that should produce a maximum rewards for the same base velocity.
The parameters of both the actor and the critic networks were initialized with those of the tracking policy.
Finally, the additional privileged inputs of the critic were set to zero as they were specific to the clip tracking task.

\begin{figure}
\centering
\vspace{0.3cm}
    \begin{subfigure}{0.235\textwidth}
        \includegraphics[width=\textwidth]{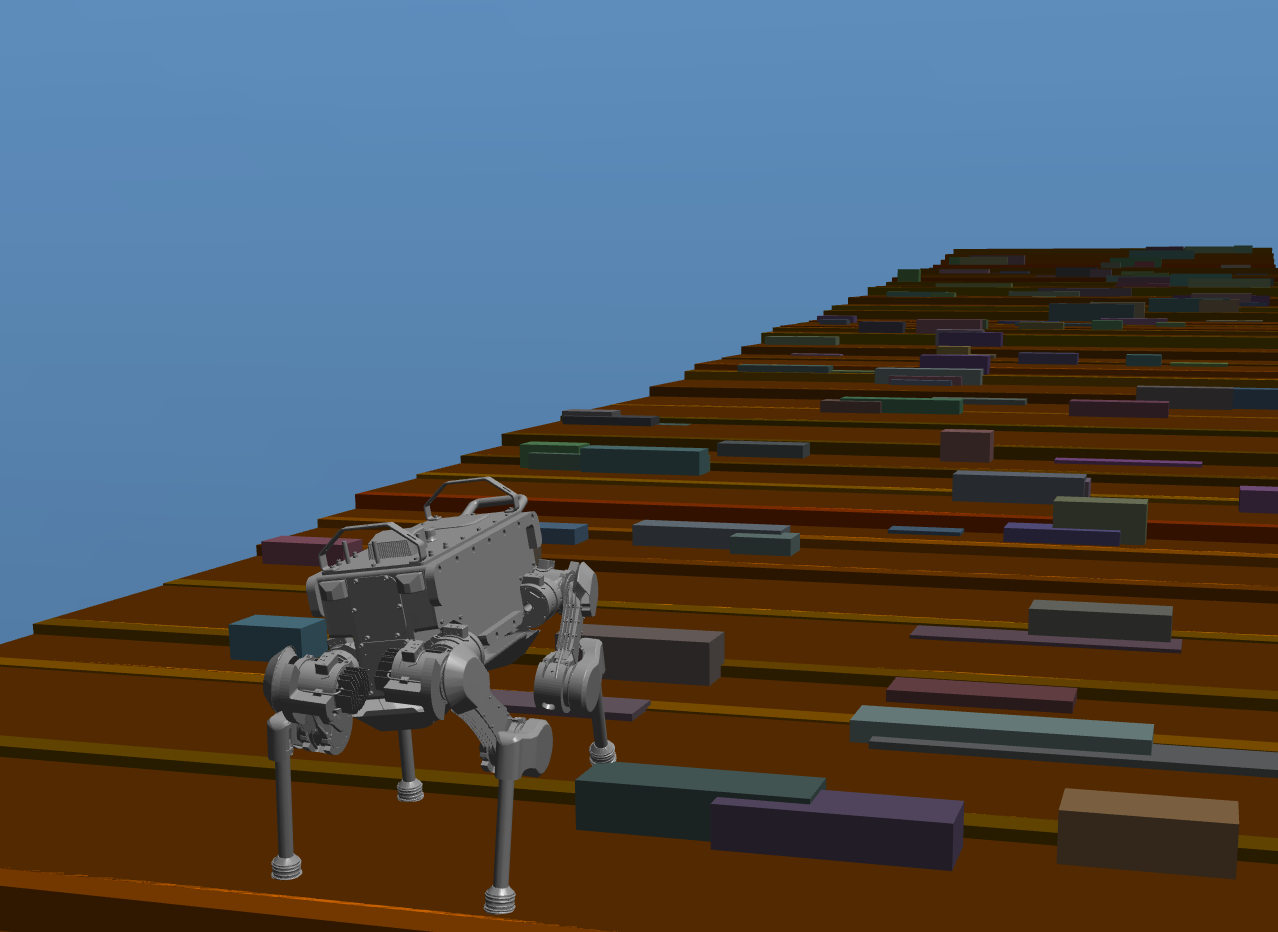}
        \caption{Stairs}
        \label{fig:stairs}
    \end{subfigure}
    \begin{subfigure}{0.235\textwidth}
        \includegraphics[width=\textwidth]{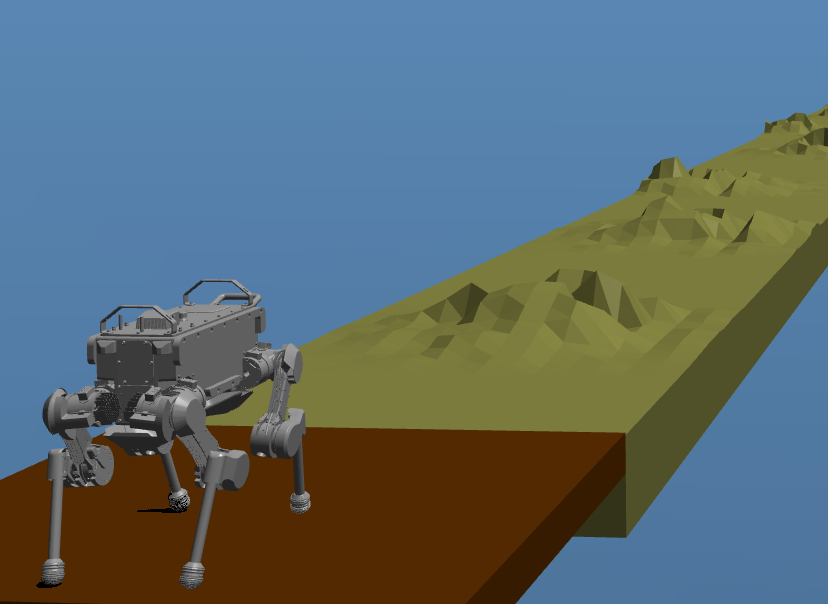}
        \caption{Procedural}
        \label{fig:terraintiles}
    \end{subfigure}
    \begin{subfigure}{0.235\textwidth}
        \includegraphics[width=\textwidth]{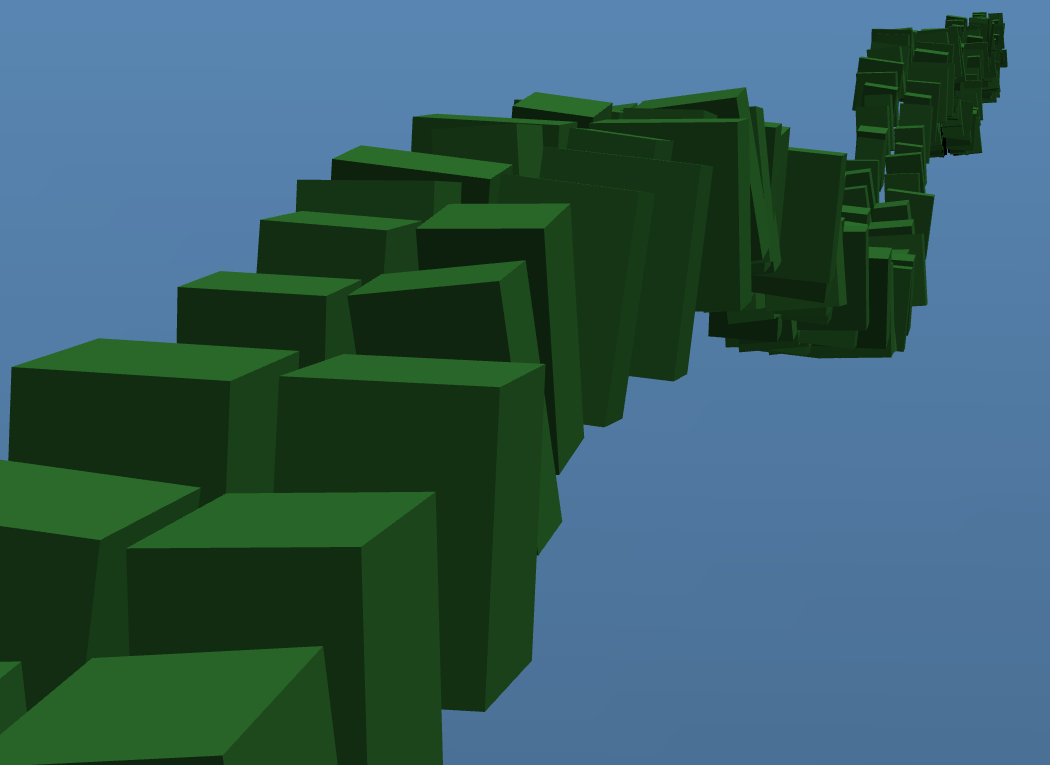}
        \caption{Wavy Steps}
        \label{fig:steppingstoneshard2}
    \end{subfigure}
    \begin{subfigure}{0.235\textwidth}
        \includegraphics[width=\textwidth]{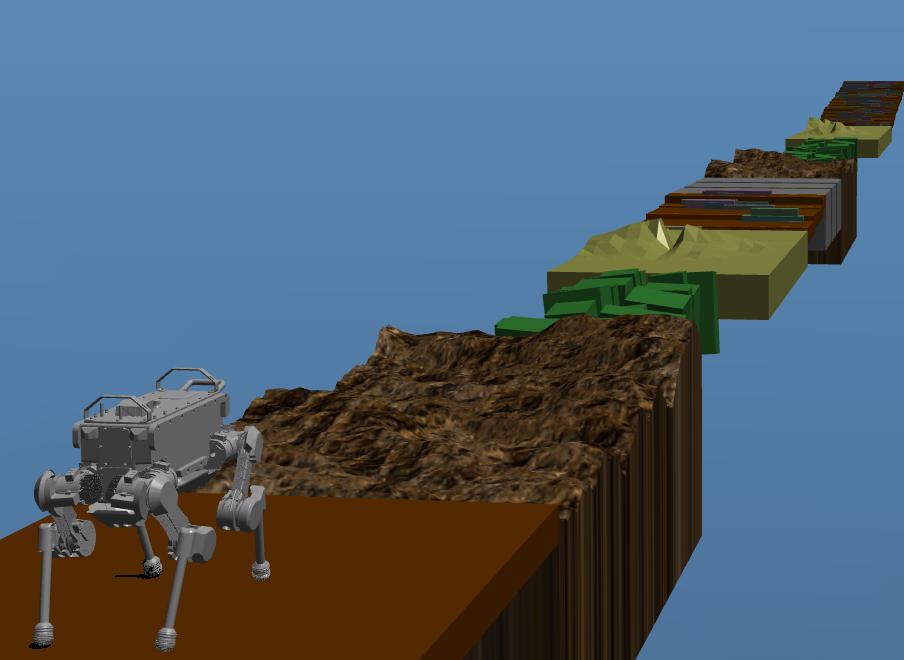}
        \caption{Mixed}
        \label{fig:combinedtiles}
    \end{subfigure}
    \caption{Evaluation and fine-tuning environments used in the experiments.}
    \label{fig:environments}
\end{figure}

\section{Experiments}
We used the Mujoco library \cite{Todorov2012mujoco} to simulate the dynamics of the full ANYmal model. The actuators were modelled with a neural network trained on real data, similar to what was done in other recent work on the ANYmal \cite{Hwangbo2019learning} to make the simulation more realistic. Our qualitiative results can be seen in the 
accompanying video.
\label{sec:result}

\subsection{Environments}
To evaluate and fine-tune our policies, we used a variety of environments, which are shown in Fig.\ \ref{fig:environments}. The \emph{stairs} environment consists of steps of randomly varying heights and randomly positioned boxes that serve as obstacles. The \emph{procedural} terrain environment contains height maps that were created by the same process as was used to generate the planner trajectories. The \emph{wavy steps} environment consists of a track of pairs of stepping stones with randomized height differences and gaps between them. The pitch and roll rotations of the stepping stones are also randomized and the overall elevation pattern follows a sine wave. Finally, we used a \emph{mixed} environment which combines the stairs, procedural terrains and wavy steps while adding two additional types of terrains: flat regions with randomly spaced and sized slits and another type of procedural terrain based on a higher resolution height map and Perlin noise.
See Appendix \ref{app:environments} for more detailed descriptions of the environments and Appendix \ref{app:moreimages} for additional screenshots in a larger format.

\subsection{Baselines}
As baselines, we ran multiple seeds \emph{from scratch} on each environment using the same architecture and learning algorithm as in the fine-tuning experiments but with randomly initialized network weights. We ran three baseline seeds for each environment, except for the wavy steps for which we observed more variance and ran ten. We report the average performance over 100 episodes of the best performing baseline seed. We also fine-tuned the baselines trained on the procedural terrains on the wavy steps track and added these policies to the seeds to take the maximum over for this environment. This was done to disentangle curriculum learning benefits of first training on procedural terrains from potential benefits due to properties of the imitated trajectories.
All baselines were trained for 20 billion environment steps but performance typically stopped improving much earlier.

\subsection{Tracking Performance and Zero-Shot Transfer}

\begin{figure}
\vspace{0.3cm}
    \centering
    \begin{subfigure}{0.43\columnwidth}
        \includegraphics[width=\columnwidth]{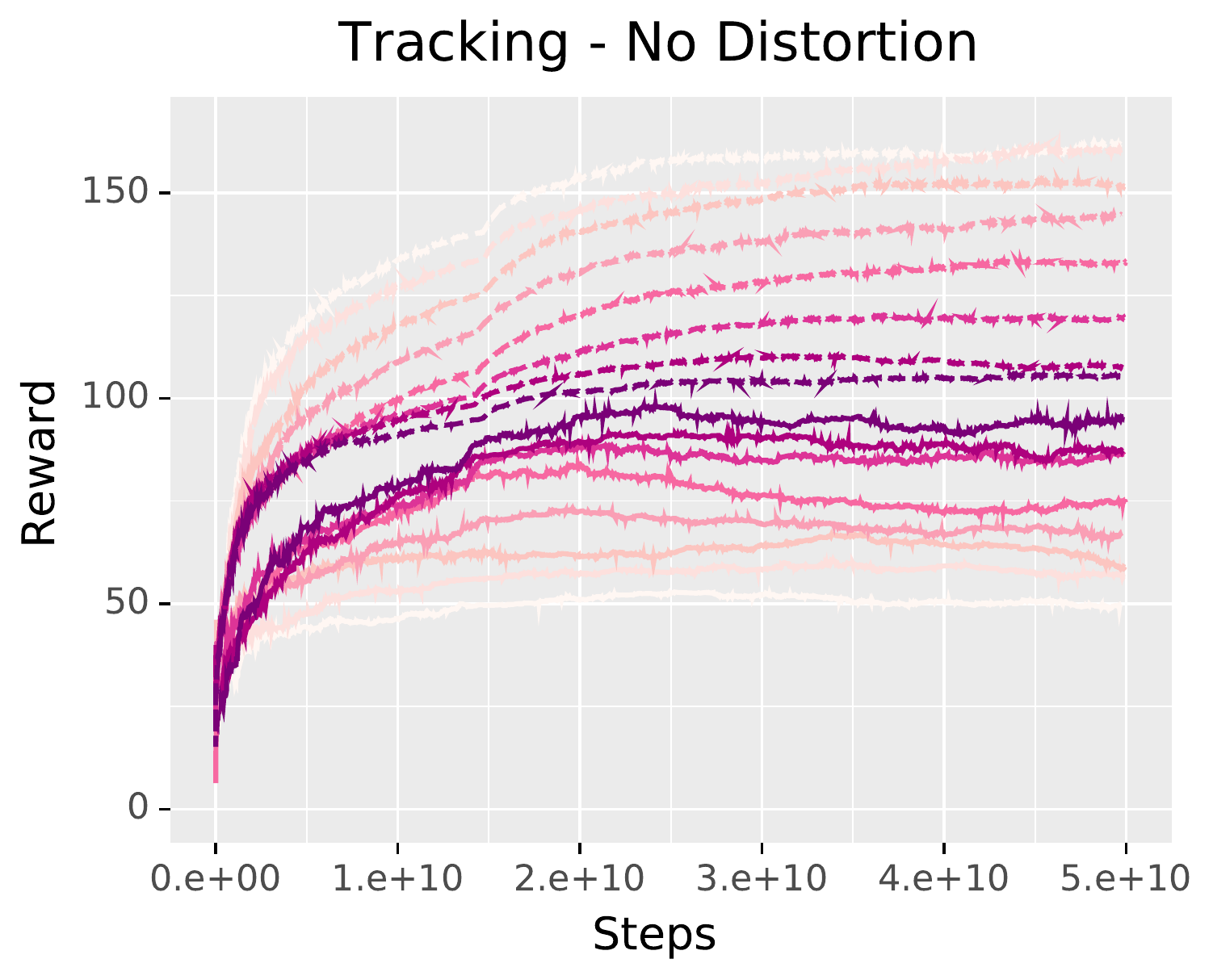}
        \caption{}
        \label{fig:tracknodist}
    \end{subfigure}
    \begin{subfigure}{0.54\columnwidth}
        \includegraphics[width=\columnwidth]{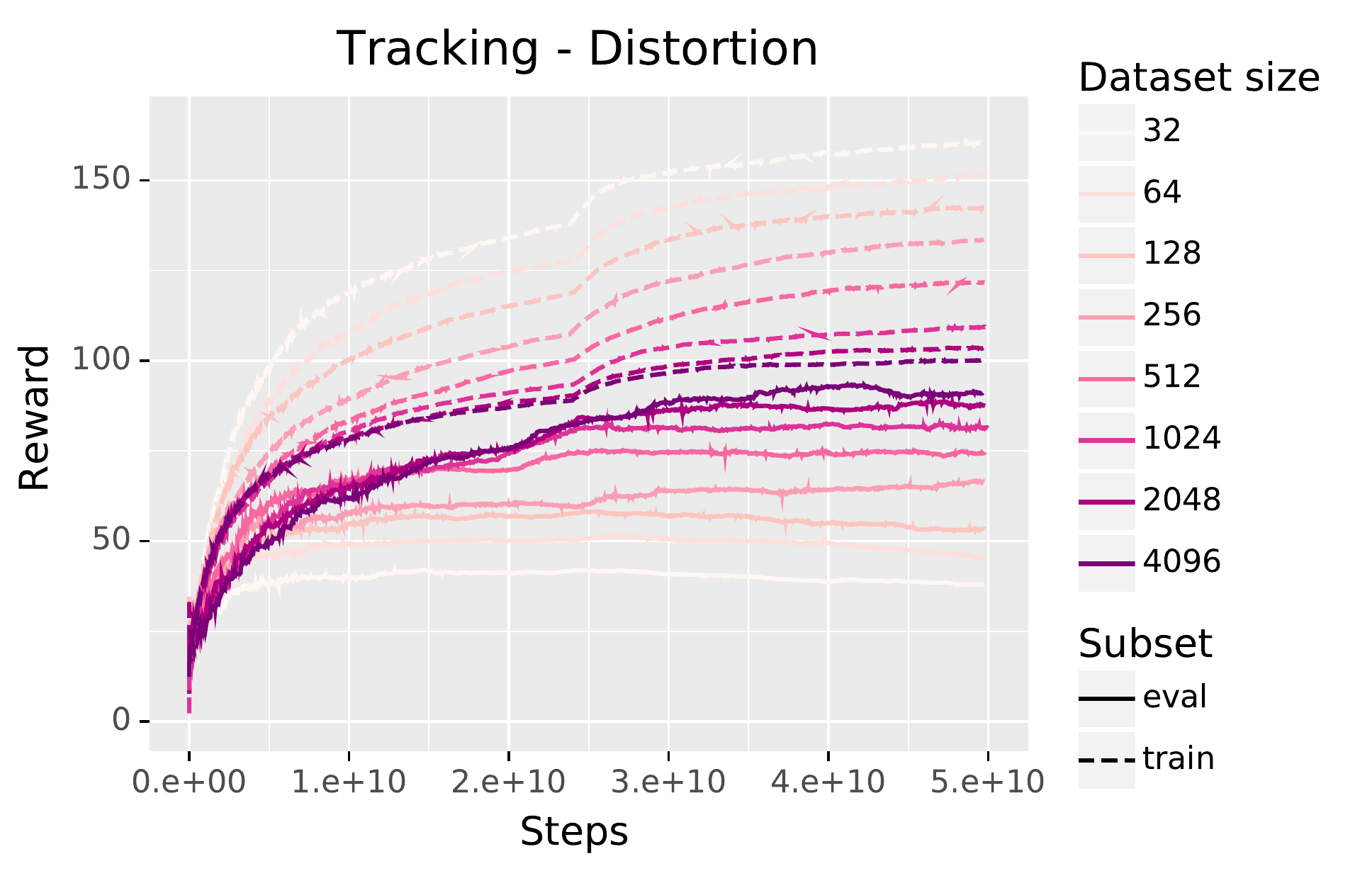}
        \caption{}
        \label{fig:trackdist}
    \end{subfigure}
    \begin{subfigure}{0.43\columnwidth}
        \includegraphics[width=\columnwidth]{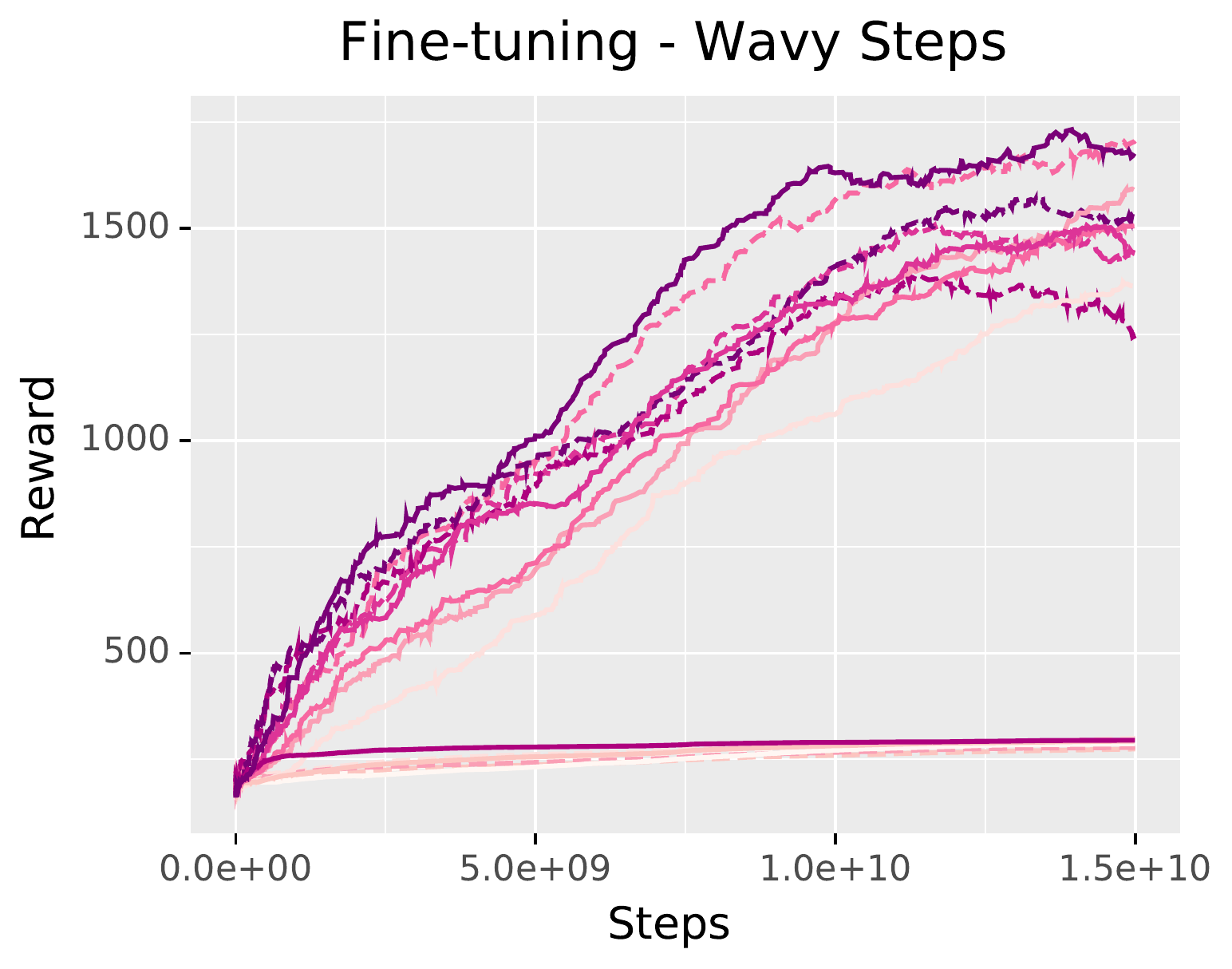}
        \caption{}
        \label{fig:stepsconv}
    \end{subfigure}
    \begin{subfigure}{0.54\columnwidth}
        \includegraphics[width=\columnwidth]{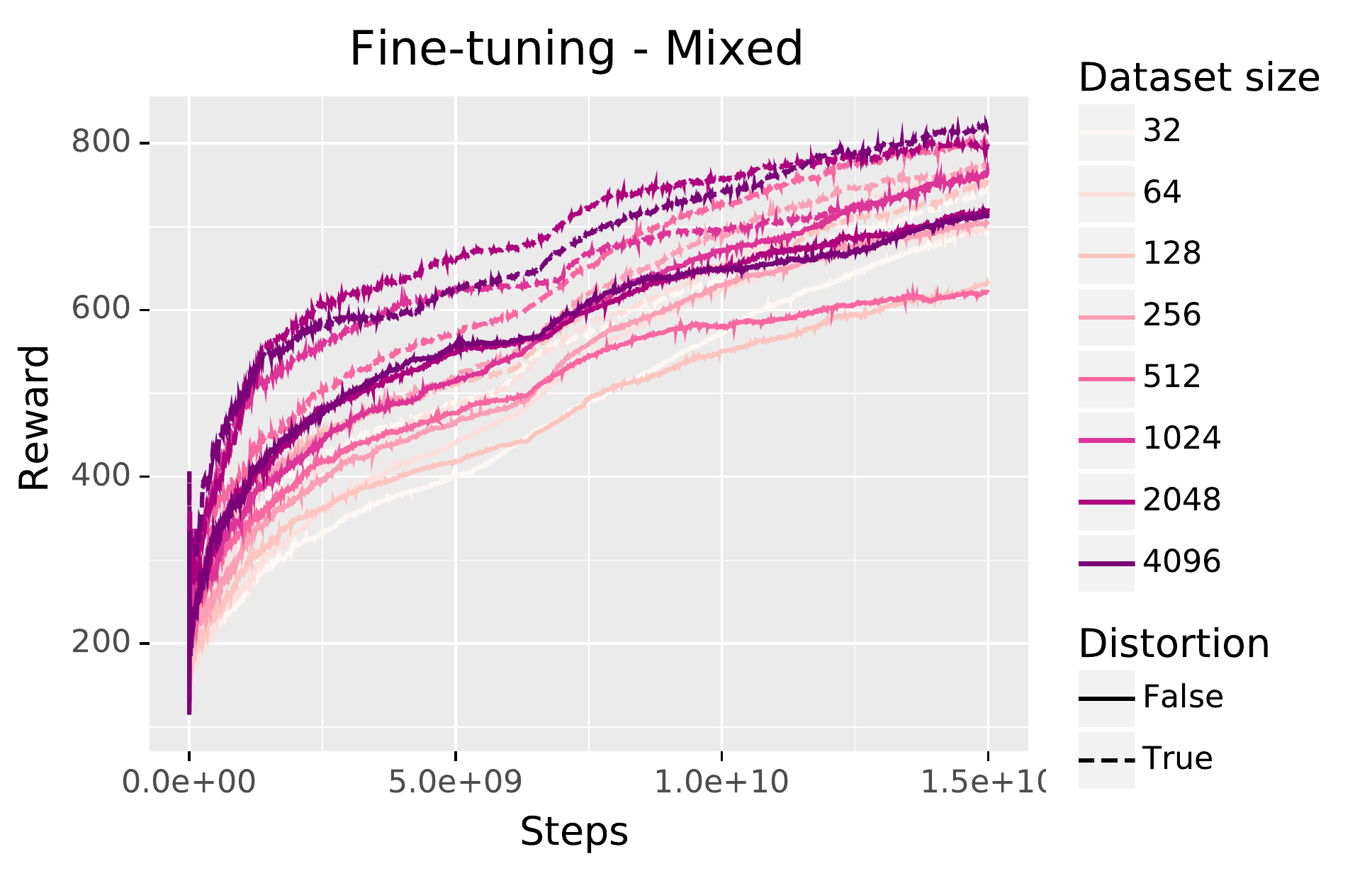}
        \caption{}
        \label{fig:mixedconv}
    \end{subfigure}
    \caption{Convergence plots for the tracking task with distortion (a) and without distortion (b) and fine-tuning convergence on the wavy steps (c) and the mixed terrains (d).}
    \label{fig:tracksizes}
\end{figure}

To find out how much data is needed to achieve good generalization, we trained tracking policies using 8 different train set sizes.
To monitor tracking performance, we used the sum of all the reward terms defined in Equations \ref{eq:r_first} to \ref{eq:r_last}.
As performance kept increasing for the distorted terrains, we trained for 50 billion steps. As is clear from Fig.\ \ref{fig:tracknodist} and Fig.\ \ref{fig:trackdist}, held-out and train performances keep approaching each other as the size of the dataset increases, showing that sufficiently many trajectories are needed for good generalization. The terrain distortions don't seem to affect this pattern.
As shown in the accompanying video, the policies trained on the full dataset can track the planned trajectories over challenging terrains that require precise foot placements.

We evaluated the tracking policies without fine-tuning on the \emph{procedural} and \emph{stairs} environments.
Episodes are $\SI{60}{s}$ long and we report the 100 episode average distance traversed in Fig.\ \ref{fig:proceduraldist} and \ref{fig:stairsdist}.
As Fig.\ \ref{fig:proceduraldist} shows, zero-shot performance on the procedural environment is better for larger datasets. The distortions seem to help a bit for the larger datasets but hurt performance for smaller sets.
The policies trained without distortions traverse about $\SI{18}{m}$ of the terrains on average. This shows, together with the tracking performance on held-out data, that the imitation learning setup can generalize to new terrains.
While this zero-shot performance is on par with the baseline (which was trained on this terrain for 20 billion steps), the behaviors of the tracking policies look qualitatively very different.
The performance on the stairs environment (not seen during training) is also directly related to the size of the datasets, as can be seen in Fig.\ \ref{fig:stairsdist} and the best performing policies are able to traverse more than $\SI{6}{m}$ on average. The task specific baseline performed much better on this task with an average of over $\SI{23}{m}$ but we consider the zero-shot performance promising.

\subsection{Fine-tuning}

\begin{figure}
\vspace{0.3cm}
    \centering
    \begin{subfigure}[c]{0.43\columnwidth}
        \includegraphics[width=\columnwidth]{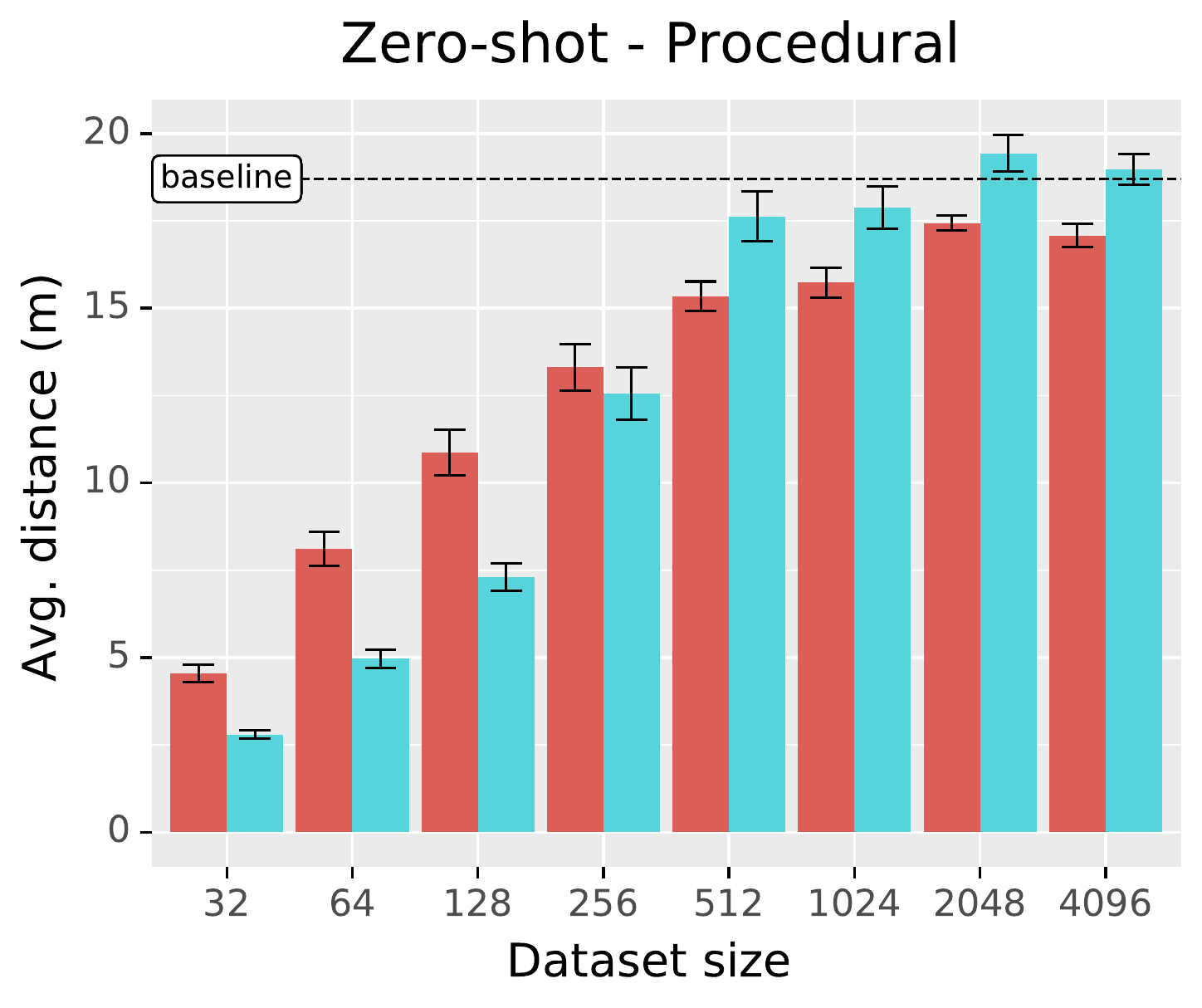}
        \caption{}
        \label{fig:proceduraldist}
    \end{subfigure}
    \begin{subfigure}[c]{0.54\columnwidth}
        \includegraphics[width=\columnwidth]{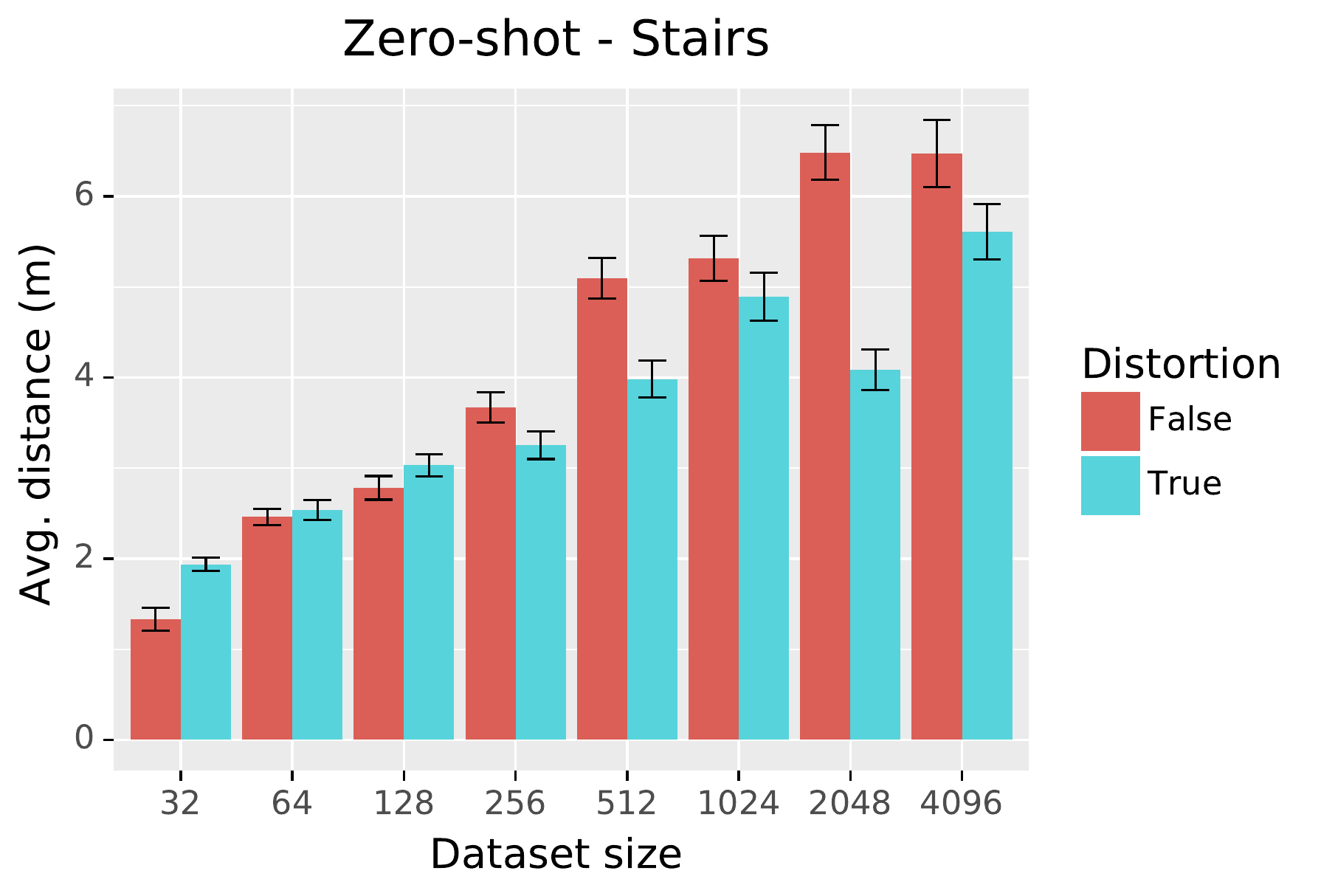}
        \caption{}
        \label{fig:stairsdist}
    \end{subfigure}
    \begin{subfigure}[c]{0.43\columnwidth}
        \includegraphics[width=\columnwidth]{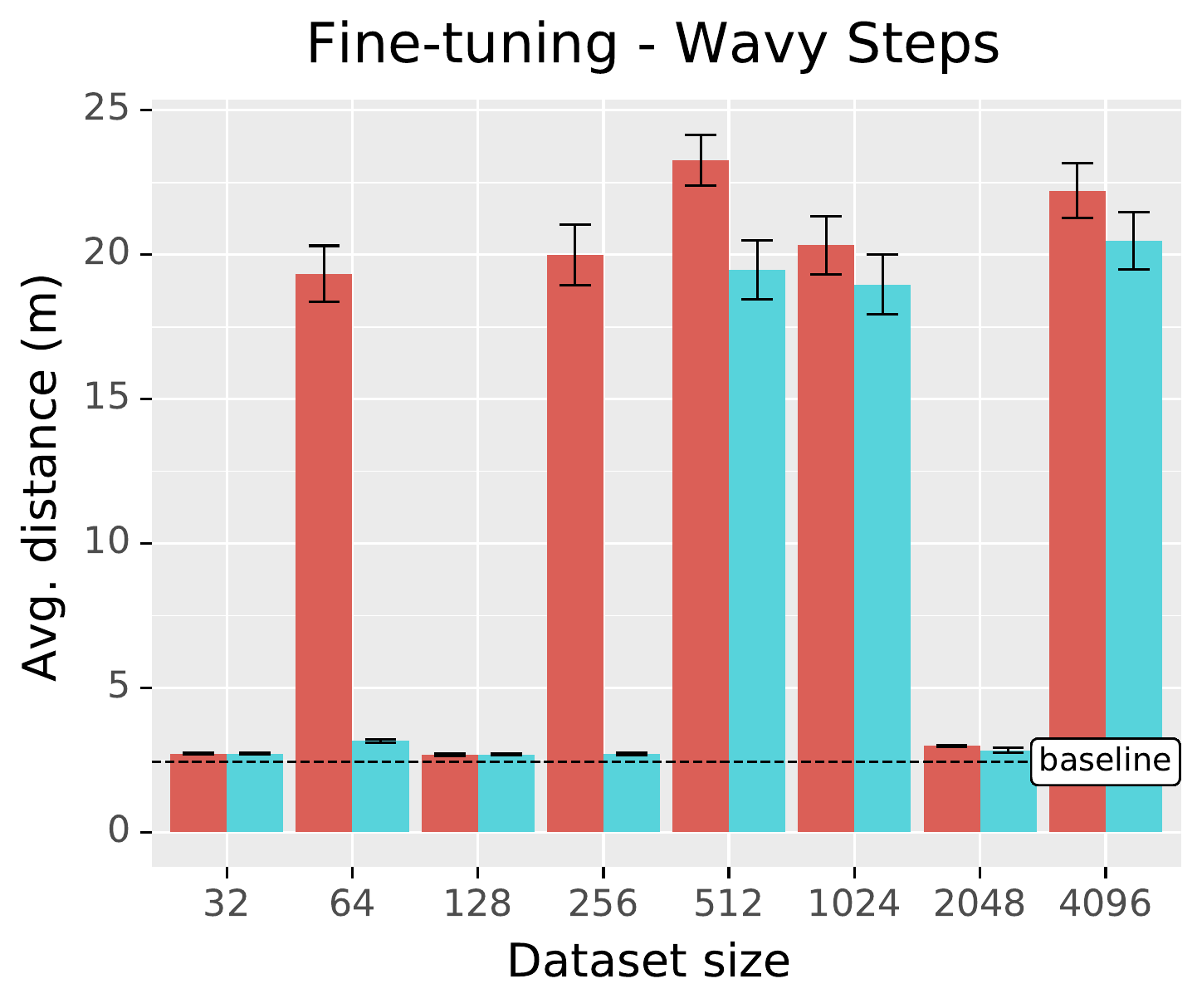}
        \caption{}
        \label{fig:wavydist}
    \end{subfigure}
    \begin{subfigure}[c]{0.54\columnwidth}
        \includegraphics[width=\columnwidth]{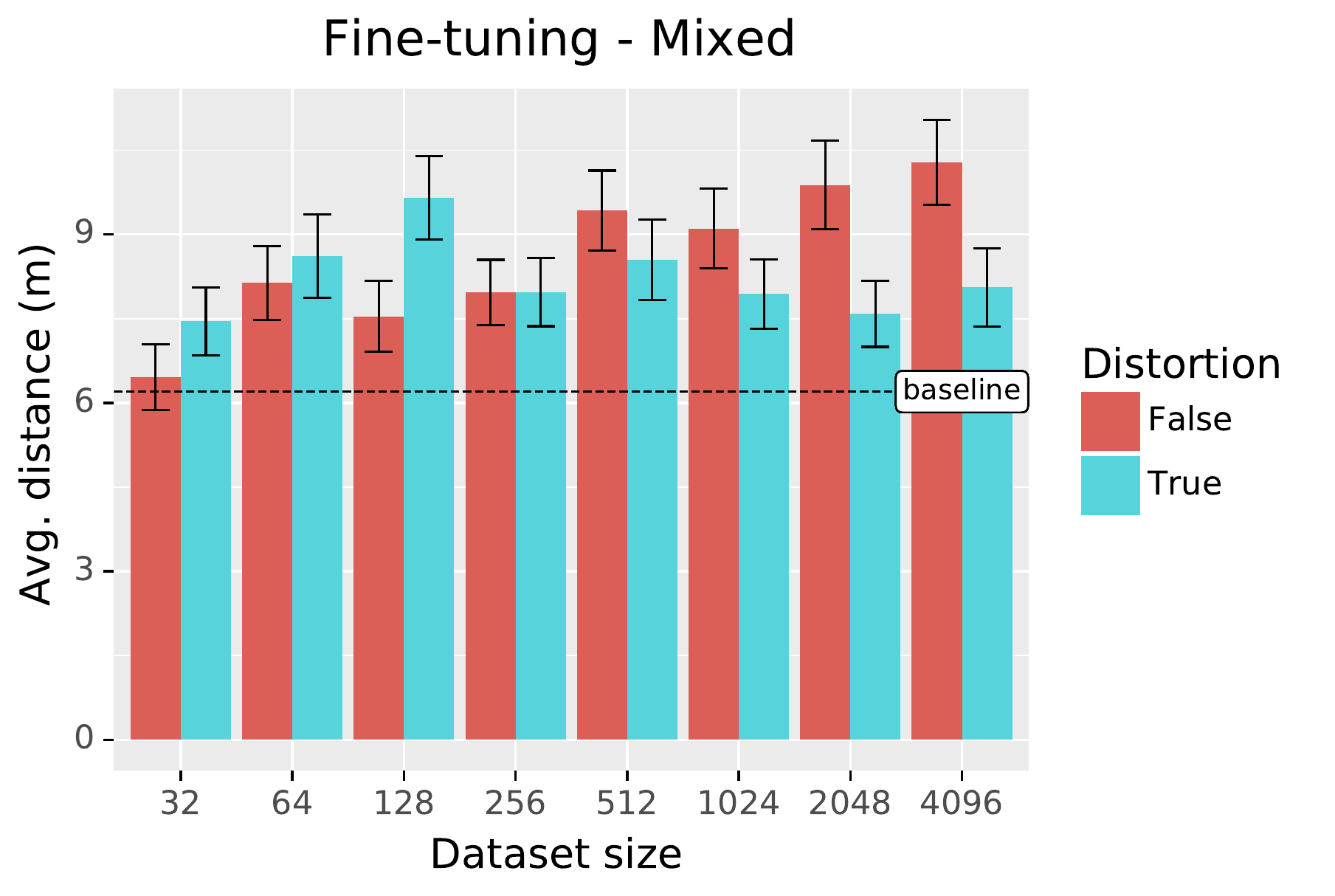}
        \caption{}
        \label{fig:mixeddist}
    \end{subfigure}
    \begin{subfigure}[c]{0.54\columnwidth}
        \includegraphics[width=\columnwidth]{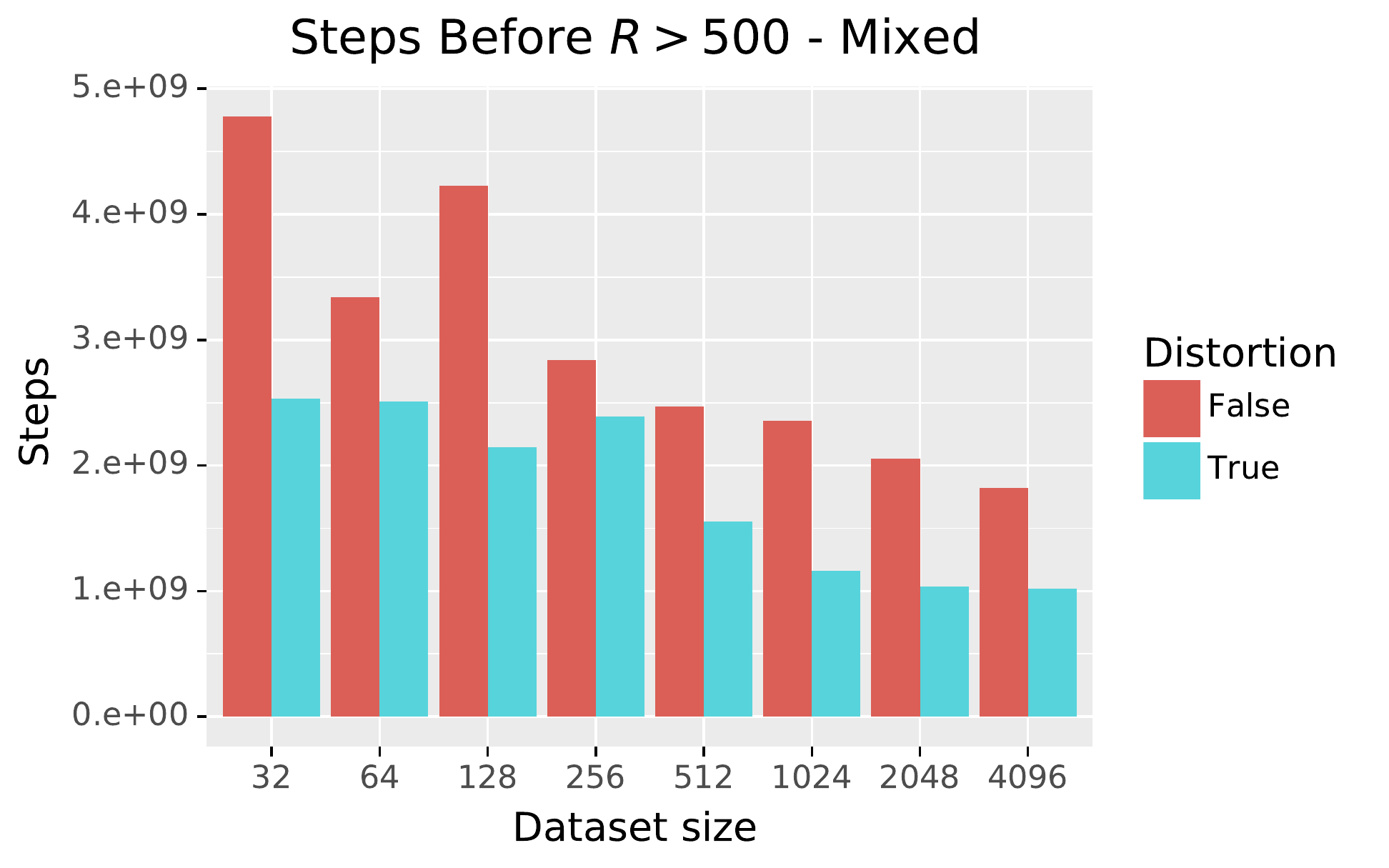}
        \caption{}
        \label{fig:stepsneeded}
    \end{subfigure}
    \begin{subfigure}[c]{0.43\columnwidth}
        \centering
        \includegraphics[scale=0.4]{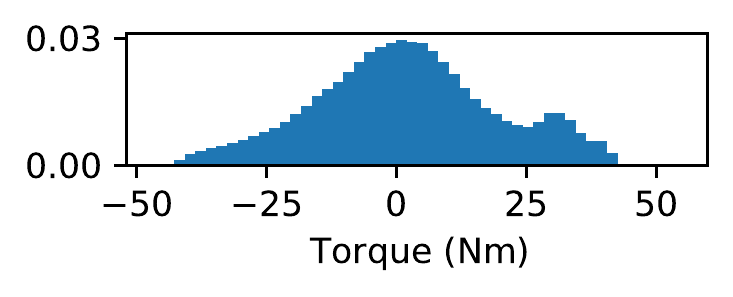}
        \includegraphics[scale=0.4]{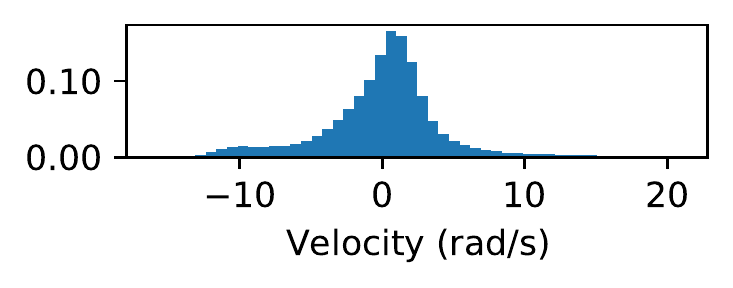}
        \caption{}
        \label{fig:torquevel}
    \end{subfigure}
    \caption{(a-d) Meters traversed for the procedural and stairs environments (zero shot) and the wavy steps and mixed terrains (after fine tuning). (d) Number of steps needed to obtain more than 500 reward on the mixed terrains (rolling average with a window size corresponding to $10\%$ of the data). (f) Torque and velocity histograms measured for a policy fine-tuned on the wavy steps terrain.}
    \label{fig:my_label}
\end{figure}

We fine-tuned the tracking policies on the wavy steps and mixed environments using 15 billion steps. The former was constructed to be hard to learn with low-level trial-and-error exploration while the latter was constructed to test generality.
Fig.\ \ref{fig:stepsconv}, \ref{fig:mixedconv} and \ref{fig:stepsneeded} show that convergence speed is generally correlated with the size of the tracking dataset.
As Fig.\ \ref{fig:wavydist} shows, the wavy steps were hard to learn and policies would either get stuck at about $\SI{2.5}{m}$ like all our baseline seeds or learn the task and traverse about $\SI{20}{m}$ on average. There seems to be a relation between dataset size and final performance. Distortions seem to hurt performance on this environment.
As Fig.\ \ref{fig:mixeddist} shows, all the tracking policies trained on all but the smallest dataset performed better than the baseline for the mixed terrains.
The effect of the distortions on final performance is not clear.
However, as shown in Fig.\ \ref{fig:stepsneeded}, the number of steps needed to fine-tune past a reward of 500 (an arbitrary number that all policies achieved at some point) was lower for the policies trained with distortions and larger datasets.
As the accompanying video shows, our policies can traverse both the wavy stepping stones parts and the rough procedural parts of the terrain. A limitation is that the gaits diverge more from the planned trajectories as fine-tuning continues.

To evaluate how promising the behaviors are for transfer to the real platform, we looked at the distributions of the joint torques and velocities. As shown in Fig.\ \ref{fig:torquevel}, there are no significant peaks above the suggested maximum torque of $\SI{40}{Nm}$ \cite{hutter2016anymal}. The largest observed torque was $\SI{54.78}{Nm}$ and only $1.28\%$ of the values were above $\SI{40}{Nm}$. The velocity distribution looks well-behaved too. The largest observed value was $\SI{21}{rad/s}$ and only $1.68\%$ of the observed values surpassed the suggested maximum of $\SI{12}{rad/s}$. We think that these values do not provide reason to suspect that the behaviors would be infeasible on the real robot.

\section{Conclusion and Discussion}

We showed that imitation from a diverse set of planned trajectories can be used to learn dynamic terrain-adaptive controllers. After fine-tuning, the policies can traverse environments that are very hard to solve with a standard RL setup.
Our data augmentation method showed mixed results but aided fine-tuning convergence speed.
While we only presented results in simulation, we used a realistic robot model and observations that can all at least be estimated on a real platform. The fine-tuning stage doesn't prevent the policies from drifting away towards gaits that don't resemble the tracking policy anymore. This can be partially remedied by early stopping but a more principled method would be to add a penalty term that enforces the policies to remain similar to the tracking policy or to the planner trajectories. A possible way to do this would be via adversarial motion priors \cite{peng2021amp}.
In future work, we also want to add domain randomization and test the controllers on a real robot via sim2real.
We hope that our method for learning dynamic terrain-adaptive controllers will allow robot platforms to move more dynamically over more challenging terrains with less engineering effort and we believe that it provides a useful starting point for research on terrain-adaptive skill reuse and adaptation.

\section*{Acknowledgment}

We'd like to thank Michael Neunert, Josh Merel, Fereshteh Sadeghi, Markus Wulfmeier, Yuval Tassa, Michael Bloesch, Saran Tunyasuvunakool, Raia Hadsell
 and Emilio Parisotto for their feedback and useful discussions. We also thank Piotr Trochim for technical support.

\bibliography{refs}  %

\begin{thebibliography}{10}
\providecommand{\url}[1]{#1}
\csname url@rmstyle\endcsname
\providecommand{\newblock}{\relax}
\providecommand{\bibinfo}[2]{#2}
\providecommand\BIBentrySTDinterwordspacing{\spaceskip=0pt\relax}
\providecommand\BIBentryALTinterwordstretchfactor{4}
\providecommand\BIBentryALTinterwordspacing{\spaceskip=\fontdimen2\font plus
\BIBentryALTinterwordstretchfactor\fontdimen3\font minus
  \fontdimen4\font\relax}
\providecommand\BIBforeignlanguage[2]{{%
\expandafter\ifx\csname l@#1\endcsname\relax
\typeout{** WARNING: IEEEtran.bst: No hyphenation pattern has been}%
\typeout{** loaded for the language `#1'. Using the pattern for}%
\typeout{** the default language instead.}%
\else
\language=\csname l@#1\endcsname
\fi
#2}}

\bibitem{Lee2020}
J.~Lee, J.~Hwangbo, L.~Wellhausen, V.~Koltun, and M.~Hutter, ``Learning
  quadrupedal locomotion over challenging terrain,'' \emph{Science robotics},
  vol.~5, no.~47, 2020.

\bibitem{Siekmann2021blind}
J.~Siekmann, K.~Green, J.~Warila, A.~Fern, and J.~Hurst, ``Blind bipedal stair
  traversal via sim-to-real reinforcement learning,'' \emph{arXiv preprint
  arXiv:2105.08328}, 2021.

\bibitem{kalakrishnan2010fast}
M.~Kalakrishnan, J.~Buchli, P.~Pastor, M.~Mistry, and S.~Schaal, ``Fast, robust
  quadruped locomotion over challenging terrain,'' in \emph{2010 IEEE
  International Conference on Robotics and Automation}.\hskip 1em plus 0.5em
  minus 0.4em\relax IEEE, 2010, pp. 2665--2670.

\bibitem{bellicoso2018dynamic}
C.~D. Bellicoso, F.~Jenelten, C.~Gehring, and M.~Hutter, ``Dynamic locomotion
  through online nonlinear motion optimization for quadrupedal robots,''
  \emph{IEEE Robotics and Automation Letters}, vol.~3, no.~3, pp. 2261--2268,
  2018.

\bibitem{Tsounis2020deepgait}
V.~Tsounis, M.~Alge, J.~Lee, F.~Farshidian, and M.~Hutter, ``Deepgait: Planning
  and control of quadrupedal gaits using deep reinforcement learning,''
  \emph{IEEE Robotics and Automation Letters}, vol.~5, no.~2, pp. 3699--3706,
  2020.

\bibitem{Carius2020mpc}
J.~Carius, F.~Farshidian, and M.~Hutter, ``Mpc-net: A first principles guided
  policy search,'' \emph{IEEE Robotics and Automation Letters}, vol.~5, no.~2,
  pp. 2897--2904, 2020.

\bibitem{Hwangbo2019learning}
J.~Hwangbo, J.~Lee, A.~Dosovitskiy, D.~Bellicoso, V.~Tsounis, V.~Koltun, and
  M.~Hutter, ``Learning agile and dynamic motor skills for legged robots,''
  \emph{Science Robotics}, vol.~4, no.~26, 2019.

\bibitem{lowrey2018plan}
K.~Lowrey, A.~Rajeswaran, S.~Kakade, E.~Todorov, and I.~Mordatch, ``Plan
  online, learn offline: Efficient learning and exploration via model-based
  control,'' \emph{arXiv preprint arXiv:1811.01848}, 2018.

\bibitem{levine2013}
\BIBentryALTinterwordspacing
S.~Levine and V.~Koltun, ``Guided policy search,'' in \emph{Proceedings of the
  30th International Conference on Machine Learning}, ser. Proceedings of
  Machine Learning Research, S.~Dasgupta and D.~McAllester, Eds., vol.~28,
  no.~3.\hskip 1em plus 0.5em minus 0.4em\relax Atlanta, Georgia, USA: PMLR,
  17--19 Jun 2013, pp. 1--9. [Online]. Available:
  \url{https://proceedings.mlr.press/v28/levine13.html}
\BIBentrySTDinterwordspacing

\bibitem{peng2020animals}
X.~B. Peng, E.~Coumans, T.~Zhang, T.-W. Lee, J.~Tan, and S.~Levine, ``Learning
  agile robotic locomotion skills by imitating animals,'' \emph{arXiv preprint
  arXiv:2004.00784}, 2020.

\bibitem{Song2019v}
H.~F. Song, A.~Abdolmaleki, J.~T. Springenberg, A.~Clark, H.~Soyer, J.~W. Rae,
  S.~Noury, A.~Ahuja, S.~Liu, D.~Tirumala, \emph{et~al.}, ``V-mpo: On-policy
  maximum a posteriori policy optimization for discrete and continuous
  control,'' \emph{arXiv preprint arXiv:1909.12238}, 2019.

\bibitem{heess2017emergence}
N.~Heess, D.~TB, S.~Sriram, J.~Lemmon, J.~Merel, G.~Wayne, Y.~Tassa, T.~Erez,
  Z.~Wang, S.~Eslami, \emph{et~al.}, ``Emergence of locomotion behaviours in
  rich environments,'' \emph{arXiv preprint arXiv:1707.02286}, 2017.

\bibitem{escontrela2020zero}
A.~Escontrela, G.~Yu, P.~Xu, A.~Iscen, and J.~Tan, ``Zero-shot terrain
  generalization for visual locomotion policies,'' \emph{arXiv preprint
  arXiv:2011.05513}, 2020.

\bibitem{shi2021reinforcement}
H.~Shi, B.~Zhou, H.~Zeng, F.~Wang, Y.~Dong, J.~Li, K.~Wang, H.~Tian, and
  M.~Q.-H. Meng, ``Reinforcement learning with evolutionary trajectory
  generator: A general approach for quadrupedal locomotion,'' \emph{arXiv
  preprint arXiv:2109.06409}, 2021.

\bibitem{xie2020allsteps}
Z.~Xie, H.~Y. Ling, N.~H. Kim, and M.~van~de Panne, ``Allsteps:
  Curriculum-driven learning of stepping stone skills,'' in \emph{Computer
  Graphics Forum}, vol.~39, no.~8.\hskip 1em plus 0.5em minus 0.4em\relax Wiley
  Online Library, 2020, pp. 213--224.

\bibitem{rudin2021learning}
N.~Rudin, D.~Hoeller, P.~Reist, and M.~Hutter, ``Learning to walk in minutes
  using massively parallel deep reinforcement learning,'' \emph{arXiv preprint
  arXiv:2109.11978}, 2021.

\bibitem{Gangapurwala2020rloc}
S.~Gangapurwala, M.~Geisert, R.~Orsolino, M.~Fallon, and I.~Havoutis, ``Rloc:
  Terrain-aware legged locomotion using reinforcement learning and optimal
  control,'' \emph{arXiv preprint arXiv:2012.03094}, 2020.

\bibitem{xie2021glide}
Z.~Xie, X.~Da, B.~Babich, A.~Garg, and M.~van~de Panne, ``Glide: Generalizable
  quadrupedal locomotion in diverse environments with a centroidal model,''
  \emph{arXiv preprint arXiv:2104.09771}, 2021.

\bibitem{gangapurwala2020guided}
S.~Gangapurwala, A.~Mitchell, and I.~Havoutis, ``Guided constrained policy
  optimization for dynamic quadrupedal robot locomotion,'' \emph{IEEE Robotics
  and Automation Letters}, vol.~5, no.~2, pp. 3642--3649, 2020.

\bibitem{Melon2020reliable}
O.~Melon, M.~Geisert, D.~Surovik, I.~Havoutis, and M.~Fallon, ``Reliable
  trajectories for dynamic quadrupeds using analytical costs and learned
  initializations,'' in \emph{2020 IEEE International Conference on Robotics
  and Automation (ICRA)}.\hskip 1em plus 0.5em minus 0.4em\relax IEEE, 2020,
  pp. 1410--1416.

\bibitem{kwon2020fast}
T.~Kwon, Y.~Lee, and M.~Van De~Panne, ``Fast and flexible multilegged
  locomotion using learned centroidal dynamics,'' \emph{ACM Transactions on
  Graphics (TOG)}, vol.~39, no.~4, pp. 46--1, 2020.

\bibitem{bogdanovic2021model}
M.~Bogdanovic, M.~Khadiv, and L.~Righetti, ``Model-free reinforcement learning
  for robust locomotion using trajectory optimization for exploration,''
  \emph{arXiv preprint arXiv:2107.06629}, 2021.

\bibitem{Peng2018deepmimic}
X.~B. Peng, P.~Abbeel, S.~Levine, and M.~van~de Panne, ``Deepmimic:
  Example-guided deep reinforcement learning of physics-based character
  skills,'' \emph{ACM Transactions on Graphics (TOG)}, vol.~37, no.~4, pp.
  1--14, 2018.

\bibitem{Merel2018neural}
J.~Merel, L.~Hasenclever, A.~Galashov, A.~Ahuja, V.~Pham, G.~Wayne, Y.~W. Teh,
  and N.~Heess, ``Neural probabilistic motor primitives for humanoid control,''
  \emph{arXiv preprint arXiv:1811.11711}, 2018.

\bibitem{Hasenclever2020comic}
\BIBentryALTinterwordspacing
L.~Hasenclever, F.~Pardo, R.~Hadsell, N.~Heess, and J.~Merel, ``Comic:
  Complementary task learning {\&} mimicry for reusable skills,'' in
  \emph{Proceedings of the 37th International Conference on Machine Learning,
  {ICML} 2020, 13-18 July 2020, Virtual Event}, ser. Proceedings of Machine
  Learning Research, vol. 119.\hskip 1em plus 0.5em minus 0.4em\relax {PMLR},
  2020, pp. 4105--4115. [Online]. Available:
  \url{http://proceedings.mlr.press/v119/hasenclever20a.html}
\BIBentrySTDinterwordspacing

\bibitem{peng2021amp}
X.~B. Peng, Z.~Ma, P.~Abbeel, S.~Levine, and A.~Kanazawa, ``Amp: Adversarial
  motion priors for stylized physics-based character control,'' \emph{arXiv
  preprint arXiv:2104.02180}, 2021.

\bibitem{Holden2017phase}
D.~Holden, T.~Komura, and J.~Saito, ``Phase-functioned neural networks for
  character control,'' \emph{ACM Transactions on Graphics (TOG)}, vol.~36,
  no.~4, pp. 1--13, 2017.

\bibitem{Winkler2018-zl}
A.~W. Winkler, C.~D. Bellicoso, M.~Hutter, and J.~Buchli, ``Gait and trajectory
  optimization for legged systems through {Phase-Based} {End-Effector}
  parameterization,'' \emph{IEEE Robotics and Automation Letters}, vol.~3,
  no.~3, pp. 1560--1567, July 2018.

\bibitem{Wachter2006}
A.~W{\"a}chter and L.~T. Biegler, ``On the implementation of an interior-point
  filter line-search algorithm for large-scale nonlinear programming,''
  \emph{Mathematical programming}, vol. 106, no.~1, pp. 25--57, 2006.

\bibitem{sen2017inverse}
M.~A. Sen, V.~Bakircioglu, and M.~Kalyoncu, ``Inverse kinematic analysis of a
  quadruped robot,'' \emph{International journal of scientific \& technology
  research}, vol.~6, no.~9, pp. 285--289, 2017.

\bibitem{Abdolmaleki2020distributional}
A.~Abdolmaleki, S.~Huang, L.~Hasenclever, M.~Neunert, F.~Song, M.~Zambelli,
  M.~Martins, N.~Heess, R.~Hadsell, and M.~Riedmiller, ``A distributional view
  on multi-objective policy optimization,'' in \emph{International Conference
  on Machine Learning}.\hskip 1em plus 0.5em minus 0.4em\relax PMLR, 2020, pp.
  11--22.

\bibitem{kingma2014adam}
D.~P. Kingma and J.~Ba, ``Adam: A method for stochastic optimization,''
  \emph{arXiv preprint arXiv:1412.6980}, 2014.

\bibitem{hochreiter1997long}
S.~Hochreiter and J.~Schmidhuber, ``Long short-term memory,'' \emph{Neural
  computation}, vol.~9, no.~8, pp. 1735--1780, 1997.

\bibitem{Todorov2012mujoco}
E.~Todorov, T.~Erez, and Y.~Tassa, ``Mujoco: A physics engine for model-based
  control,'' in \emph{2012 IEEE/RSJ International Conference on Intelligent
  Robots and Systems}.\hskip 1em plus 0.5em minus 0.4em\relax IEEE, 2012, pp.
  5026--5033.

\bibitem{hutter2016anymal}
M.~Hutter, C.~Gehring, D.~Jud, A.~Lauber, C.~D. Bellicoso, V.~Tsounis,
  J.~Hwangbo, K.~Bodie, P.~Fankhauser, M.~Bloesch, \emph{et~al.}, ``Anymal-a
  highly mobile and dynamic quadrupedal robot,'' in \emph{2016 IEEE/RSJ
  international conference on intelligent robots and systems (IROS)}.\hskip 1em
  plus 0.5em minus 0.4em\relax IEEE, 2016, pp. 38--44.

\bibitem{clevert2015fast}
D.-A. Clevert, T.~Unterthiner, and S.~Hochreiter, ``Fast and accurate deep
  network learning by exponential linear units (elus),'' \emph{arXiv preprint
  arXiv:1511.07289}, 2015.

\bibitem{nair2010rectified}
V.~Nair and G.~E. Hinton, ``Rectified linear units improve restricted boltzmann
  machines,'' in \emph{Icml}, 2010.

\end{thebibliography}
\bibliographystyle{IEEEtran}  %

\clearpage
\appendices
\section{Terrain Generation Procedure}
\label{app:terrain}
For each terrain, we first generated an $8\times 8$ matrix of coefficients sampled uniformly from $[0, 0.2]$ and took the Kronecker product with a $2\times 2$ matrix of ones to create a pattern of tiles. Subsequently, a random shift was applied to the columns to prevent the tiles from always being at the same locations. Finally, three $x-y$ locations were sampled uniformly from the region covered by the height map, which served as centers $\mathbf{c}_i$ of radial basis functions $f_i(\mathbf{p})=\exp(-\|\mathbf{c}_i-\mathbf{p}\|/h_i)$, where $\mathbf{p}$ is a point on the $xy$-plane and $h_i$ is a bandwidth parameter which was chosen as $h_i=\exp(-2\eta_i)$, where $\eta_i$ is uniformly sampled from $[0, 1]$. These radial basis functions were used to scale down all the values in the terrain grid based on their distance of the centers. 
The height fields were scaled down to always have a maximum height of $\SI{30}{cm}$ and represent an area of $\SI{2}{m}\times\SI{2}{m}$.

\section{Environment Details}
\label{app:environments}

\subsection{Procedural}
A series of height fields generated with the procedure in Sec.\ \ref{app:terrain} are generated (15 in total) and connected together by platforms of $\SI{3}{m}\times \SI{2}{m}$.

\subsection{Stairs}
The step heights were uniformly sampled between $\SI{0}{cm}$ and $\SI{10}{cm}$. The step lengths were sampled between $\SI{10}{cm}$ and $\SI{75}{cm}$. The steps were $\SI{2}{m}$ wide.
Each step also had three boxes that with heights sampled from the same distribution as the steps heights. The horizontal positions of the boxes were sampled to be within some maximum distance of each other that was sampled from $(\SI{0}{m},\SI{1}{m}0)$ every episode. This would cause the boxes to be somewhat bundled together. Whenever two boxes overlapped with each other, their positioning along the length of the stairs ($x$-axis) was offset by $\SI{5}{cm}$.

\subsection{Wavy Steps}

The \emph{wavy steps} environment was constructed out of a series of pairs of boxes for the robot to step on. To make the task more challenging, various properties of the track were randomized. All distributions over continuous intervals were uniform. All units are centimeters. 
The spaces separating each pair of steps laterally were sampled from $(10, 15)$, the vertical offset and longitudinal offset of each step from $(-5, 5)$, the length of the gaps in the longitudinal direction from $(15, 20)$ and the length and the width of the steps of each pair from $(20, 40)$ and $(30, 60)$ respectively. The steps were $\SI{72}{cm}$ high and we applied random rotations on the pitch and roll axes with values sampled from $(-0.15, 0.15)$ radians. Note that these rotations could significantly widen the gaps between the steps. Finally, the function $\sin((x-6)/3)$ was added to the heights of the steps, where $x$ is the longitudinal distance in meters.

\subsection{Mixed}

The mixed terrains contained five segments which where randomly sampled from to create different terrain combinations at each episode. The procedural and stairs terrains were identical to the descriptions above. A procedural segment corresponded to a single terrain segment of $\SI{4}{m}$. A stairs segment corresponded to five steps. The third terrain segment was $\SI{4}{m}$ of the wavy steps track but without the sine function being added to the heights. The forth terrain was a segment of $\SI{3}{m}$ of flat platforms separated by slits/gaps. The platform lengths were sampled from $(\SI{0.1}{m},\SI{1}{m})$ and the lengths of the gaps from $(\SI{0.15}{m},\SI{0.2}{m})$.
Finally the fifth terrain was a $\SI{4}{m}$ long height map with Perlin noise (6 harmonics) of a maximum height of $\SI{0.5}{m}$ and a resolution of 65 vertices per meter. All segments were $\SI{2}{m}$ wide and the total track combined 15 of them.

\section{Additional Images of the Environments}
\label{app:moreimages}

\begin{figure*}
\centering
\vspace{0.3cm}
    \begin{subfigure}{0.31\textwidth}
        \includegraphics[width=\textwidth]{stairsblue}
        \caption{Stairs}
    \end{subfigure}
    \begin{subfigure}{0.31\textwidth}
        \includegraphics[width=\textwidth]{terraintiles}
        \caption{Procedural}
    \end{subfigure}
    \begin{subfigure}{0.31\textwidth}
        \includegraphics[width=\textwidth]{steppingstoneshardsnake}
        \caption{Wavy Steps}
    \end{subfigure}
    \begin{subfigure}{0.31\textwidth}
        \includegraphics[width=\textwidth]{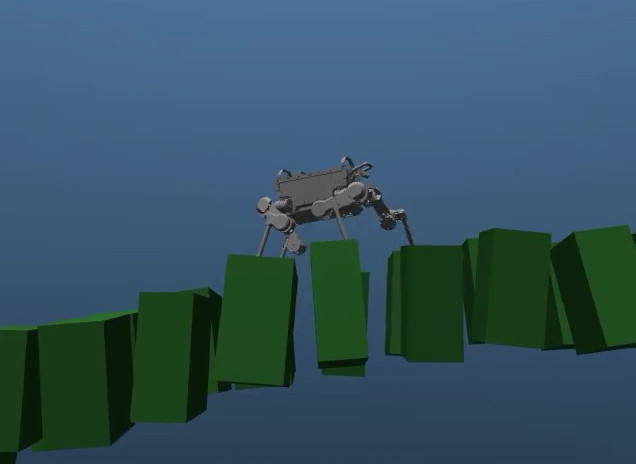}
        \caption{Wavy Steps (side view)}
    \end{subfigure}
    \begin{subfigure}{0.31\textwidth}
        \includegraphics[width=\textwidth]{combinedtiles}
        \caption{Mixed}
    \end{subfigure}
    \begin{subfigure}{0.31\textwidth}
        \includegraphics[width=\textwidth]{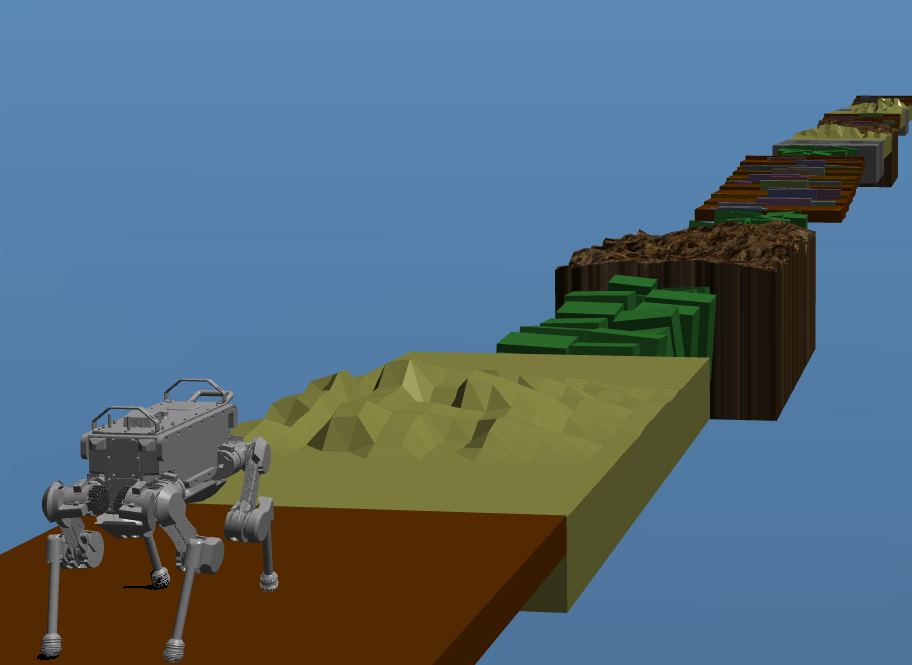}
        \caption{Mixed (different sample)}
    \end{subfigure}
    \caption{Additional and larger images of the environments used in the experiments.}
    \label{fig:moreenvironments}
\end{figure*}

See Fig.\ \ref{fig:moreenvironments}.

\section{Learning Algorithm Hyperparameters}
\label{app:hypers}
See Tabel \ref{tab:hypers} for an overview of the hyperparameter settings used in the experiments.
We refer to \cite{Abdolmaleki2020distributional} and \cite{Song2019v} for the details of the MO-VMPO algorithm but will explain the relevant hyperparameters here.
The MO-VMPO algorithm consists of an E-step and M-step which both optimize objectives under KL constraints using Lagrange multipliers. 
The E-step constraints are defined for each of the $k$ objectives via hyperparameters $\epsilon_l$ and corresponding Lagrange multpliers $\eta_l$. The M-step only has a single KL-divergence score but since this is a KL-divergence between Gaussian distributions, the mean and covariance terms can be separated and are constrained using their own individual hyperparameter  and lagrange multiplier pairs $(\epsilon_{\mu}, \alpha_{\mu})$ and $(\epsilon_{\Sigma}, \alpha_{\Sigma})$, respectively.
The Lagrange multipliers are updated automatically but their initial values can strongly influence learning and are therefore relevant hyperparameters as well.

\begin{table}[]
    \caption{Learning hyperparameters.}
    \centering
    \begin{tabular}{lc}
    \toprule \midrule
    \textbf{Hyperparameter} & \textbf{Value} \\
    \midrule
    training &\\
    \hspace{2em} Adam learning rate   & $10^{-4}$\\
    \hspace{2em} target network update period & $100$\\
    \hspace{2em} discount        & $0.99$ \\
    \hspace{2em} batch size      & $1024$ \\
    \midrule
    MO-VMPO &\\
    \hspace{2em}  variational distribution KL constraints, $\epsilon_{l}$    & $0.002$ \\
    \hspace{2em}  initial variational dist.\ Lagrange multipliers, $\eta_{l,0}$          & $10$\\ 
    \hspace{2em}  mean KL constraint, $\epsilon_{\mu}$    & $0.1$ \\
    \hspace{2em}  covariance KL constraint, $\epsilon_{\Sigma}$    & $10^{-5}$ \\
    \hspace{2em}  initial mean lagrange multiplier, $\alpha_{\mu,0}$ & $0.1$\\
    \hspace{2em}  initial covariance Lagrange multiplier, $\alpha_{\Sigma,0}$ & $100$\\
    \bottomrule
    \end{tabular}
    \label{tab:hypers}
\end{table}

\section{Architecture Details}
\label{app:architecture}
The actor is a conditional normal distribution parameterized by a deep neural network.
Since the observations don't provide full state information, we added memory to the network by incorporating an LSTM \cite{hochreiter1997long}. More precisely, the local terrain information $\mathbf{M}(t)$ was first processed by two convolutional layers without pooling with respectively $16$ filters of dimension $5\times 5$ and $32$ filters of dimension $3\times 3$. Each layer was followed by an ELU non-linearity \cite{clevert2015fast}. The final outputs were passed through a fully connected layer of 128 dimensions and concatenated with the remaining observation features. Subsequently, the concatenated features were passed through a 256 unit fully connected tanh layer, followed by the 512 unit large LSTM. The LSTM output was passed through by one more 512 unit fully connected tanh layer and finally projected to the parameters of a multivariate normal distribution with diagonal covariance.

The critic received the same observations as the actor but also an embedding vector representing the clip number and a set of future reference body positions and quaternions $\{\xi_{i}(t+k)\colon i\in \mathcal{B},k\in\{2,4,6,8,10\}\}$, where $\mathcal{B}$ is the set of body indices. The terrain and proprioceptive features went through the same computations as in the actor (but with separate parameters) up to the concatenation of the image and proprioceptive paths. This was followed by three fully connected 
 layers (two tanh and one ReLU \cite{nair2010rectified}, an unintentional inconsistency) and finally a linear transformation to each of the value predictions.
The critic parameters were kept during fine-tuning but it's additional observations (see Sec. \ref{sec:obs}) were replaced with zeros.

\section{Comparison of Preserving or Resetting the Critic Parameters}
\label{app:reloading}
\begin{figure}
\centering
    \begin{subfigure}{0.48\columnwidth}
        \includegraphics[width=\columnwidth]{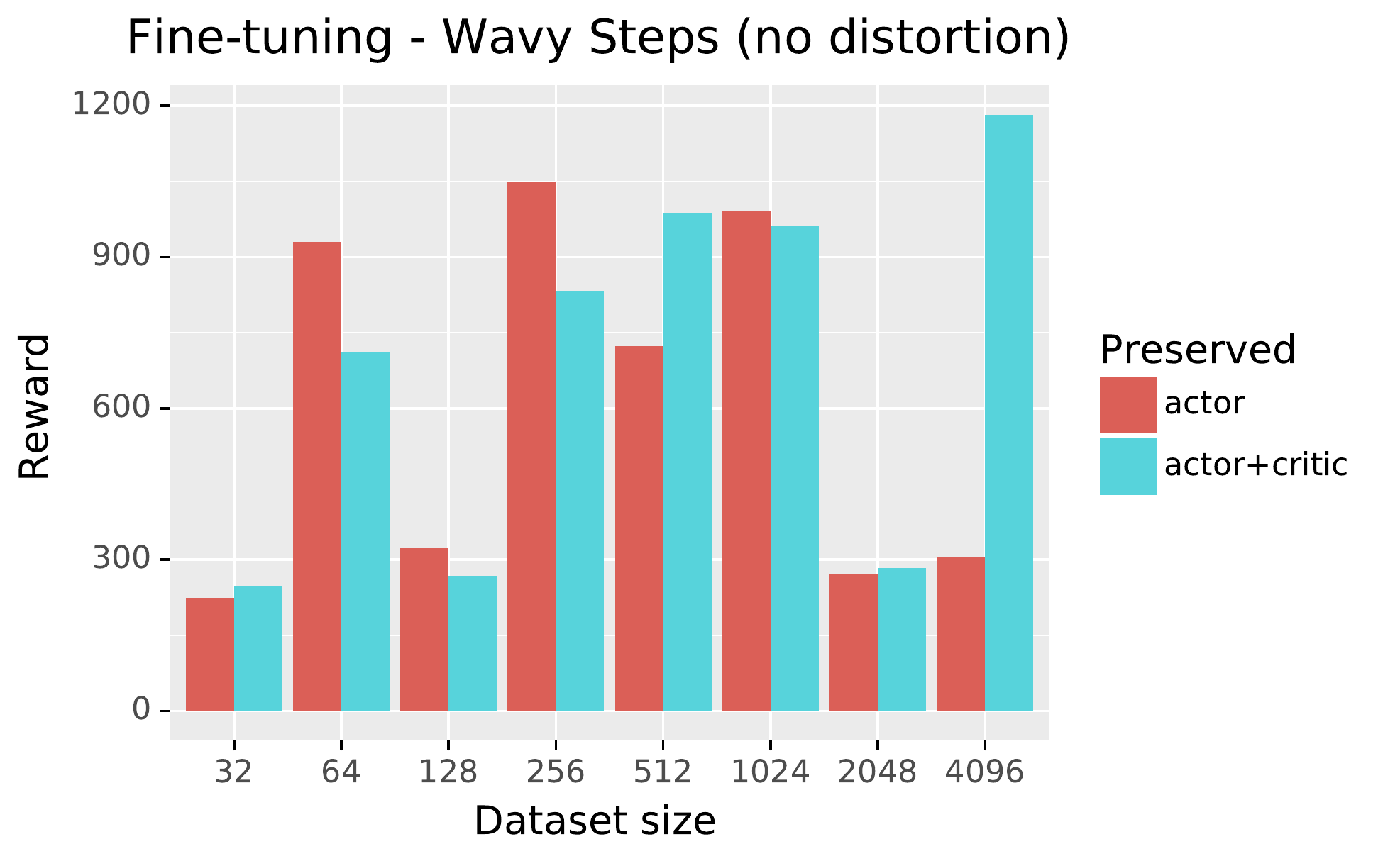}
        \caption{}
    \end{subfigure}
    \begin{subfigure}{0.48\columnwidth}
        \includegraphics[width=\columnwidth]{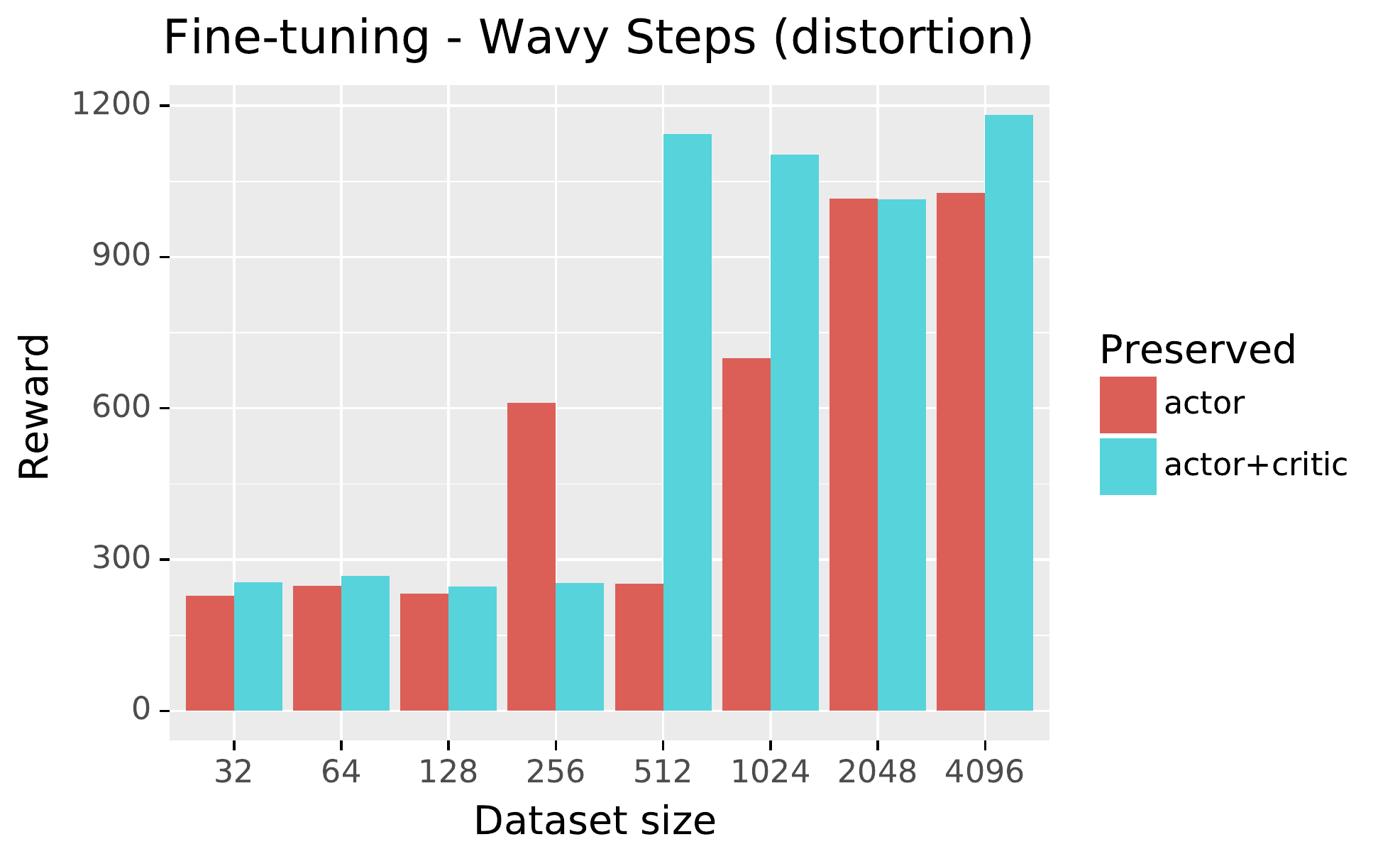}
        \caption{}
    \end{subfigure}
    \caption{Comparisons of the average rewards over the first 8 billion steps for fine-tuning on wavy steps when preserving both the actor and critic weights or only those of the actor.}
    \label{fig:reloading}
\end{figure}

To see if resetting the critic instead of preserving it would lead to substantially different results, we ran an additional fine-tuning experiment on the wavy steps task. 
As can be seen in Fig.\ \ref{fig:reloading}, the overall pattern of performance looks the same for both preserving or resetting the critic weights. The most notable difference is that without distortions, the run in which the critic was reset failed on the largest dataset in the sweep. We suspect that this is just an unlucky seed and that more runs are needed to verify this result. However, overall, these results don't suggest that rerunning the full experiments while resetting the critic would have lead to substantially different results.

\end{document}